%% file: dualRE-main.tex
\documentclass[sigconf]{acmart}
\usepackage{hhline}

\usepackage{subfigure}
\usepackage{multirow}
\usepackage{color, verbatim,caption}
\usepackage{hyperref}
\usepackage{enumitem}
\usepackage{dsfont}
\usepackage{balance}

\usepackage[ruled ]{algorithm2e}
\usepackage{array}
\newcolumntype{L}[1]{>{\raggedright\let\newline\\\arraybackslash\hspace{0pt}}m{#1}}
\newcolumntype{C}[1]{>{\centering\let\newline\\\arraybackslash\hspace{0pt}}m{#1}}
\newcolumntype{R}[1]{>{\raggedleft\let\newline\\\arraybackslash\hspace{0pt}}m{#1}}

\usepackage{booktabs} 
\usepackage{bbm}

\setcopyright{iw3c2w3}

\usepackage[font=small,labelfont=bf,textfont=md]{caption}

\acmDOI{10.475/123_4}
\acmISBN{123-4567-24-567/08/06}

\acmConference[WWW'19]{The Web Conference}{May 13--17 2019}{San Francisco, CA, USA}
\acmYear{2019}
\copyrightyear{2019}

\acmArticle{4}
\acmPrice{15.00}

\editor{Jennifer B. Sartor}
\editor{Theo D'Hondt}
\editor{Wolfgang De Meuter}

\begin{document}
\title{Learning Dual Retrieval Module for Semi-supervised \\Relation Extraction}

\author{Hongtao Lin}
\affiliation{%
  \institution{University of Southern California}
  \city{Los Angeles}
  \state{California}
  \postcode{90007}
}
\email{lin498@usc.edu}

\author{Jun Yan}
\affiliation{%
  \institution{Tsinghua University}
  \streetaddress{30 Shuangqing Rd}
  \city{Haidian Qu}
  \state{Beijing Shi}
  \country{China}
}
\email{j-yan15@mails.tsinghua.edu.cn}

\author{Meng Qu}
\affiliation{%
  \institution{University of Illinois at Urbana-Champaign}
  \city{Champaign}
  \state{Illinois}
  \postcode{61820}
}
\email{mengqu2@illinois.edu}

\author{Xiang Ren}
\affiliation{%
  \institution{University of Southern California}
  \city{Los Angeles}
  \state{California}
  \postcode{90007}
}
\email{xiangren@usc.edu}

\renewcommand{\shortauthors}{Lin et al.}

\begin{abstract}
Relation extraction is an important task in structuring content of text data, and becomes especially challenging when learning with weak supervision---where only a limited number of labeled sentences are given and a large number of unlabeled sentences are available. Most existing work exploits unlabeled data based on the ideas of self-training (i.e., bootstrapping a model) and multi-view learning (e.g., ensembling multiple model variants). However, these methods either suffer from the issue of semantic drift, or do not fully capture the problem characteristics of relation extraction. In this paper, we leverage a key insight that \textit{retrieving sentences expressing a relation} is a dual task of \textit{predicting relation label for a given sentence}---two tasks are complementary to each other and can be optimized jointly for mutual enhancement. To model this intuition, we propose DualRE, a principled framework that introduces a retrieval module which is jointly trained with the original relation prediction module. In this way, high-quality samples selected by retrieval module from unlabeled data can be used to improve prediction module, and vice versa. Experimental results\footnote{\small Code and data can be found at \url{https://github.com/ink-usc/DualRE}.} on two public datasets as well as case studies demonstrate the effectiveness of the  DualRE approach. 
\end{abstract}


\maketitle

\input{1-introduction.tex}
\input{3-definition.tex}
\input{4-model.tex}
\input{5-experiment.tex}
\input{2-related-work.tex}
\input{6-conclusion.tex}

\bibliographystyle{ACM-Reference-Format}


\end{document}

%% file: 1-introduction.tex
\section{Introduction}

\begin{figure}
	\centering
	\includegraphics[width=0.47\textwidth]{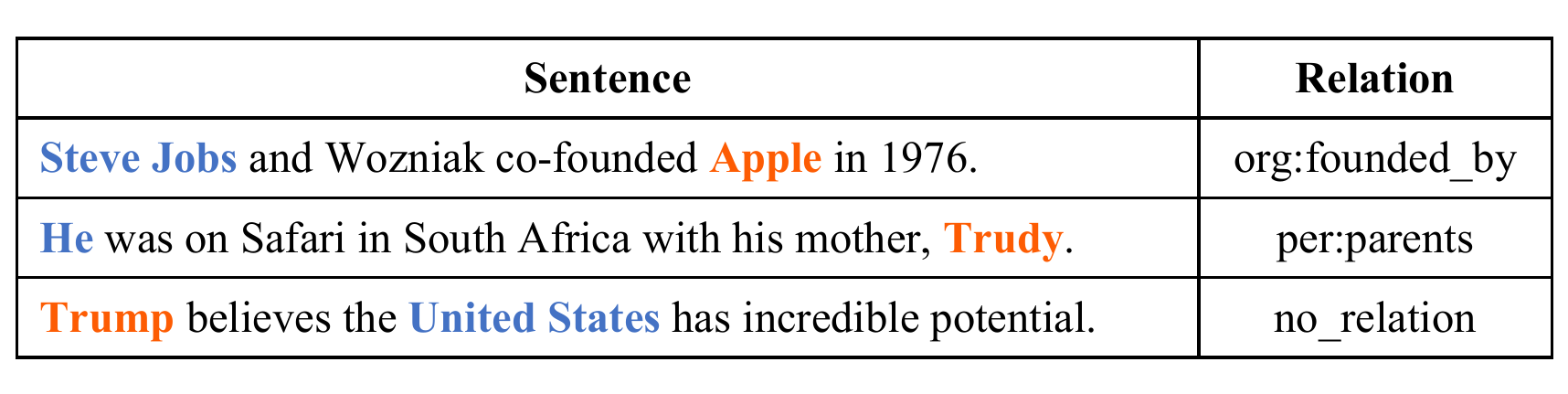}
	\caption{\textbf{Sentence-level Relation Extraction Task}. This task concerns about predicting relation of two entities mentioned in a sentence. Tokens in orange and blue are subject and object entities in correspondence.}
	\label{fig::re_intro}
	\vspace{-0.5cm}
\end{figure}

Relation extraction (RE) plays a key role in turning massive, unstructured text corpora into structures of factual knowledge, and has a wide range of downstream applications in many domains~\cite{lin2016neural,ren2017cotype,toutanova2015representing}. Given two entities in a sentence, the goal of (sentence-level) relation extraction is to predict the relation expressed between two entities based on their sentence context. For example, in Fig.~\ref{fig::re_intro}, we aim to classify each sentence into several relation labels, such as ``\textsf{org:founded\_by}'', ``\textsf{per:parents}'', and ``\textsf{no\_relation}''. While many work have been done on developing supervised relation extraction models~\cite{zelenko2003kernel,bunescu2005shortest,mooney2006subsequence,zeng2017incorporating,zhang2017position}, they typically require a large amount of human-annotated data (i.e., sentences that are manually assigned with the appropriate relation labels) for model training. Despite its wide applications, the process of manually annotating a large training set with many relation labels is too expensive and error-prone. The rapid emergence of large, domain-specific text corpora (e.g., social media content, customer reviews, scientific publications) calls for methods that can automatically extract entity relationships from text with \textit{less human supervision}.

In recent years there has been a surge of interest in applying \textit{distant supervision} to relation extraction~\cite{mintz2009distant, zeng2015distant,lin2016neural,zeng2017incorporating}. The general idea is to leverage facts stored in external knowledge bases (KBs) to annotate sentences in an unlabeled corpus as training data---two entities co-occurring in a sentence will be labeled with their KB relations regardless of the specific context.
Although distant supervision automates the process of labeling sentences, it suffers from two major limitations. First, not all sentences where two entities co-occur are expressing their KB relations (i.e., context-agnostic label noise). 
Second, in some application domains the target relations may not be found in existing knowledge bases (e.g., WikiData), making it hard to apply distant supervision for label generation. 
It is thus desirable to develop methods that can rely on less human-annotated data for training relation extraction models.
This motivates the study of \textit{semi-supervised relation extraction} methods~\cite{brin1998extracting, agichtein2000snowball,sun2012active}, which seek to leverage a limited number of labeled sentences and a large unlabeled (background) corpus for model training. Such setting has broad use cases, including cold-start knowledge base construction\footnote{\small\url{https://tac.nist.gov/2017/KBP/ColdStart/index.html}} and question answering~\cite{fader2014open}.

In this paper, we study the problem of semi-supervised relation extraction. Towards this goal, there are broadly two kinds of methods, i.e., \textit{self-ensembling} methods, and \textit{self-training} methods. The main idea behind self-ensembling methods~\cite{rasmus2015semi, tarvainen2017mean, french2017self, miyato2018virtual} is to impose the assumption of ``consistency under perturbation''---the prediction on the unlabeled instances should remain unchanged under small perturbations either on the data, or on the model parameters. 
In contrast, self-training methods~\cite{paass1993assessing, rosenberg2005semi} aim to incrementally promote high-confidence predictions over unlabeled instances into labeled training set, and improve the model by re-training with the updated labeled set. However, these methods suffer from the limitations as follows.
(1) \emph{Insufficient Supervision}: 
The process of enforcing prediction consistency between model variants in self-ensembling methods does not involve promoting new labeled instances from the unlabeled set. This makes it hard to improve models when ``consistency-based'' supervision is relatively weak compared to that from actual instances.
(2) \emph{Semantic Drift}~\cite{curran2007minimising}: While the self-training methods annotates unlabeled sentences as extra training data, the incremental promotion process can be biased (i.e., provides incorrect predictions), and may accumulate errors (i.e., including falsely-labeled instances into training set). 
To overcome aforementioned challenges, we aim to design an approach that can generate high-quality labeled instances (as additional supervision) from the unlabeled corpus in an effective manner.

\begin{figure}
\vspace{-0.2cm}
	\centering
	\includegraphics[width=0.47\textwidth]{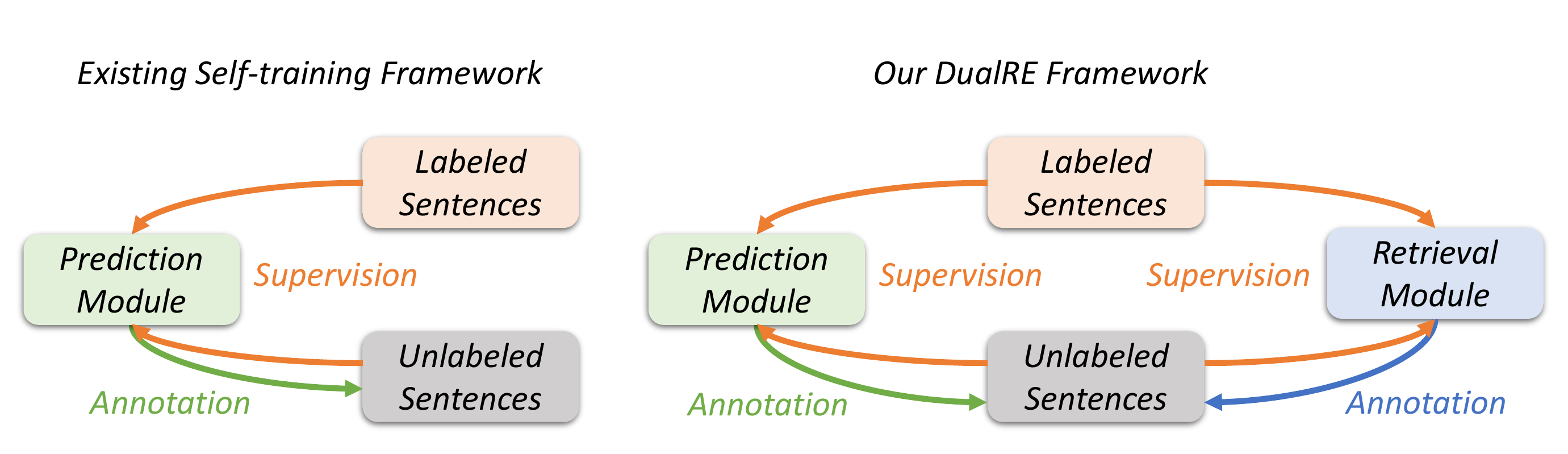}
	\vspace{-0.1cm}
	\caption{\textbf{Illustration of the differences between the proposed DualRE framework and existing Self-training framework.} \textbf{Left}: Self-training framework uses a single prediction module to annotate unlabeled sentences for generating additional labeled data. \textbf{Right}: DualRE framework leverages a prediction module and a retrieval modules for generating labeled data, where the newly labeled sentences are used to improve both modules. Two modules are jointly optimized and mutually enhance each other.}
	\label{fig::framework}
	\vspace{-0.4cm}
\end{figure}

Intuitively, \textit{retrieving sentences for a given relation} can be seen as a dual problem of \textit{predicting the relation expressed in a sentence} (see the right-hand side of Fig.~\ref{fig::framework} for illustration).
Models for the retrieval problem can help obtain more relevant sentences to a given relation from an auxiliary unlabeled corpus, for improving models for the prediction problem; while models for the prediction problem can predict the relation labels of the unlabeled sentences, and therefore help determine the quality of relevant sentences to collect (under each relation) for benefiting retrieval model. Such an observation of ``problem duality" motivates us to solve the primal and dual problems jointly, and leverage the duality to overcome the aforementioned challenges.
Specifically, as the prediction and retrieval models annotate/retrieve unlabeled sentences from the auxiliary corpus, both models provide additional training data to each other and thus help resolve the insufficient supervision issue. Meanwhile, during the joint learning process, both models provide feedback to correct the output of each other, helping generate high-quality labeled data and resolve semantic drift issue. 

To model the above ideas, we propose a novel framework, called \textbf{DualRE}, for semi-supervised relation extraction. DualRE consists of a prediction module and a retrieval module (see Fig.~\ref{fig::model} for illustration of the model design). Given a sentence $x$, the prediction module aims to determine its relation label $y$ with a neural relation extraction model by approximating $p(y|x)$ (e.g., position-aware LSTM~\cite{zhang2017position}). Meanwhile, the retrieval module employs a neural learning-to-rank model to approximate $p(x|y)$ (e.g., using pointwise or pairwise loss) to facilitate ranking sentences $\{x\}$ to identify relevant ones for a query relation $y$. During training, the two modules are jointly optimized to provide feedback for each other, by alternating between two processes as follows. In the first process, we fix the prediction module, and use additional training data promoted by both modules to update the retrieval module. Similarly, in the second process, the retrieval module is fixed, and we generate extra training data using both modules and update the prediction module. By doing so, the retrieval module can consistently provide additional supervision to improve the prediction module. As the two modules are jointly leveraged to promote new training data (via reference the label distribution), it enables the prediction and retrieval module to mutually correct each other and generate high-quality annotation. 

We conduct extensive experiments on two public relation extraction datasets to evaluate DualRE\footnote{Source code: \url{https://github.com/INK-USC/DualRE}}. Experimental results show that DualRE consistently outperforms several state-of-the-art methods under different data settings (by varying the size of labeled data and unlabeled data). Detailed case study demonstrates that DualRE is able to select quality instances to guide the model training. 

In summary, in this paper we make the following contributions:
\begin{itemize}[leftmargin=0.5cm,noitemsep,nolistsep]
    \item We propose a novel framework for semi-supervised relation extraction, which encourages the (primal) prediction task and the (dual) retrieval task to mutually benefit each other by generating additional training instances from an unlabeled corpus.
    \item We develop a joint learning algorithm to alternatively optimize the prediction and retrieval modules, such that they can collaborate with each other and improve the quality of annotations.
    \item We conduct extensive experiments on two public datasets. Experimental results and case studies validate the effectiveness of the proposed framework.
\end{itemize}

%% file: 3-definition.tex
\section{Background and Problem}
In this section, we introduce the relevant concepts and notations, and formally define semi-supervised relation extraction task.

\smallskip
\noindent \textbf{Entity Mention.}
An entity mention $e$ is a token span (or word sequence) that refers to a real-world entity. In Fig.~\ref{fig::re_intro}, the word ``Apple'' is an entity mention that maps to the company \textit{Apple Inc.}. 

\smallskip
\noindent \textbf{Relation Mention and Label.}
A relation mention consists of two entity mentions and the sentence $x$ they co-occur in (i.e., the two entity mentions are sub-strings of the sentence). Here we consider entity relations as directional connections between two entities. Therefore, we denote the left entity as subject entity $e_s = (x_{s_1}, ..., x_{s_p})$ and the right entity as object entity $e_o = (x_{o_1}, ..., x_{o_q})$, where $s_i$ and $o_j$ are the token indexes of subject and object entities in the sentence, respectively. 
A relation mention $(x, e_s, e_o)$ can be assigned with a relation label $r \in \mathcal{R}$ (i.e., $e_s$ and $e_o$ express relation $r$ in the sentence $x$). We use $\mathcal{R}$ to denote the pre-defined set of relations of interests in the downstream application. In particular, undefined or irrelevant relations are assigned with a special label ``\textsf{no\_relation}''.


\smallskip
\noindent \textbf{Problem Definition.}
In this paper, our goal is to learn a model to effectively predict the relation label between two entities $e_s$ and  $e_o$ for a relation mention $(x, e_s, e_o)$. 
Fig.~\ref{fig::re_intro} presents some examples for the task. Each row demonstrates a relation mention with two entities highlighted in blue (subject entity) and red (object entity). For relation mentions that are not expressing target relations of interests (or do not express any relation) like the third example, we assign ``\textsf{no\_relation}'' as a special label in the relation label set. Moreover, the third example also demonstrates the difference between \textit{sentence-level} and \textit{instance-level} relation extraction. Despite the well-known fact that \textit{Donald Trump} is the president of \textit{United States}, which can be extracted from multiple sentences in the corpus, we cannot infer such relation from this specific sentence and thus should return a label that is most appropriate for the given sentence. 
Formally, we define the task of sentence-level relation extraction is defined as follows.
\begin{definition}
\label{def::re}
\textbf{(Sentence-level Relation Extraction)}
Given a sentence $x = (x_1, x_2, ..., x_n)$, a subject entity $e_s$ and an object entity $e_o$ in the sentence, the task of sentence-level relation extraction is to predict the relation label $r \in \mathcal{R}$ given the sentence and two entities, where $\mathcal{R}$ is a pre-defined set of relations and the \textsf{no\_relation}.
\end{definition}

Under semi-supervised learning setting, we aim to learn the relation extraction model using both \textit{labeled relation mentions} (which are manually annotated) and \textit{unlabeled relation mentions} (of which entity mentions are identified by some existing named entity recognition tools). Formally, let $\mathbf{x} = (x, e_s, e_o)$ denote a relation mention and $y$ denote its corresponding relation label $r$ from the set $\mathcal{R}$. We define the task of semi-supervised relation extraction as follows.

\begin{definition}
\label{def::semi}
\textbf{(Semi-supervised Relation Extraction)} Given a set of labeled relation mentions $L = \{(\mathbf{x}_i, y_i)\}_{i=1}^{N_L}$ and a set of unlabeled relation mentions $U = \{\mathbf{x}_j\}_{j=1}^{N_U}$,  semi-supervised relation extraction task aims to learn a multi-class classification model $f$ that (1) fits the labeled training data $L$; and (2) captures the information in the unlabeled data $U$, at the same time.
\end{definition}

In the following sections, we stick with the denotation of $\mathbf{x}$ and $y$ when referring to the relation mention $(x, e_o, e_s)$ and relation $r$.

\smallskip
\noindent \textbf{Our Focus.}
In this work, we assume entity mentions have been recognized from text using existing tools. In our experiments, the entity mentions are provided by the public datasets~\cite{hendrickx2009semeval,zhang2017position}. Similarly, the set of relations $\mathcal{R}$ are also given by the datasets. In particular, we focus on the cases that the labeled instances are manually annotated with reliable labels, as compared to the noisy labels obtained via distant supervision~\cite{mintz2009distant}. Here we do not consider applying distant supervision to generate pseudo labels over the unlabeled relation mentions, and leave this as future work.

%% file: 4-model.tex
\section{The \textsf{DualRE} Framework}
\label{sec::dualre_framework}

\begin{figure}[!t]
	\centering
	\includegraphics[width=0.42\textwidth]{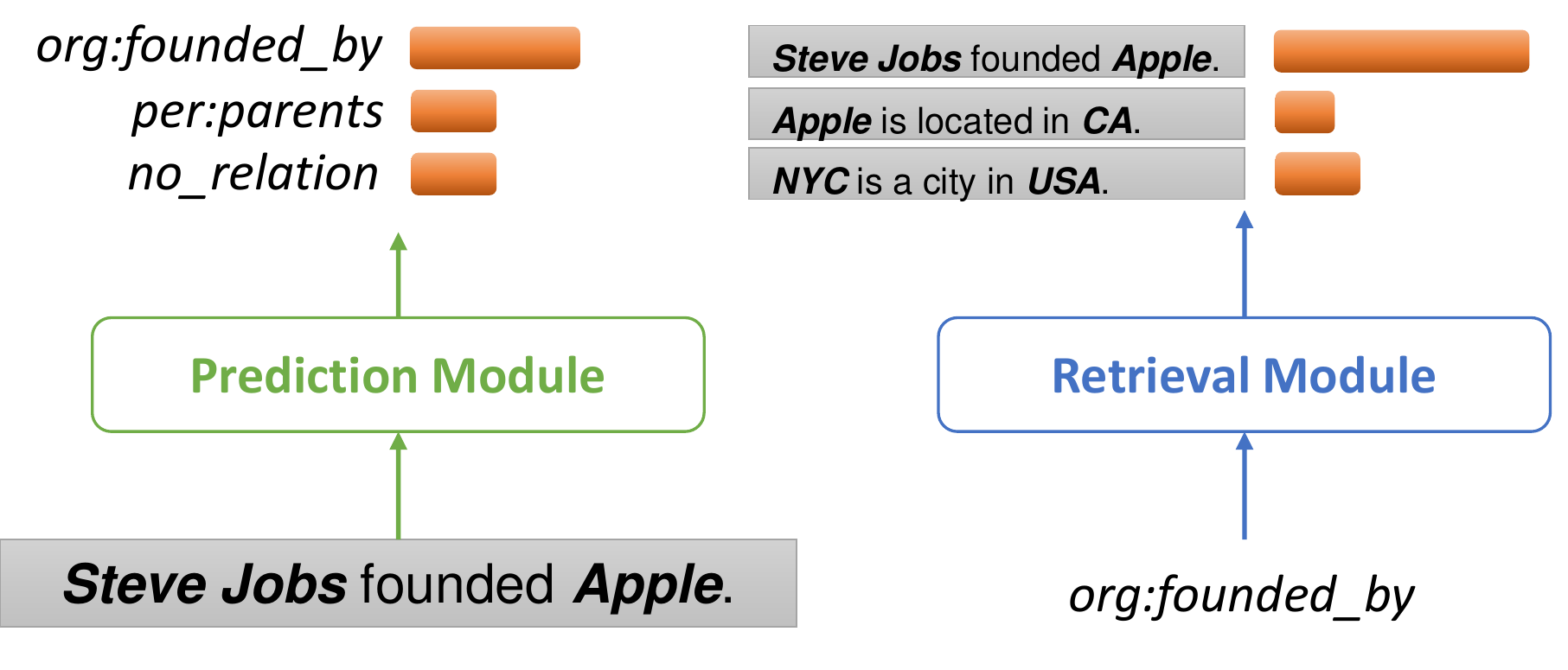}
	\caption{\textbf{Illustration of the Two Modules in DualRE.} \textbf{Left}: The prediction module focuses on the primal task, that is, predicting the relation label $y$ of a sentence $x$. \textbf{Right}: The retrieval module focuses on the dual task, which aims at retrieving relevant sentences $x$ for a query relation $y$.}
	\label{fig::model}
	\vspace{-0.2cm}
\end{figure}

\smallskip
\noindent \textbf{Framework Overview.}
This section introduces our dual learning approach to semi-supervised relation extraction. 
Given that \textit{retrieving sentences expressing a relation} can be seen as a dual task of \textit{predicting the relation expressed in a sentence}, we propose DualRE to take advantage of the dual property of two tasks and overcome the challenges mentioned in the introduction. At a high level, DualRE trains a module for the prediction task and a module for the retrieval task jointly. The retrieval module helps obtain relevant sentences for each relation from an auxiliary unlabeled corpus, which are then used as additional training data for the prediction module. Meanwhile, the prediction module predicts the relation labels for the unlabeled sentences, and determines the relation labels for retrieved sentences, which then serve as training data to improve the retrieval module. Two modules are jointly optimized to mutually enhance each other, generating additional, high-quality labeled data to address the problems of insufficient supervision and semantic drift.

Formally, our framework consists of a prediction module $\mathcal{P}_\theta$ and a retrieval module $\mathcal{Q}_\phi$, where $\theta$ and $\phi$ are their model parameters. The prediction module focuses on the primal task, i.e., predicting the relation label for a given relation mention. Therefore, for a mention-label pair $(\mathbf{x},y)$, it models the conditional probability $p_\theta(y|\mathbf{x})$. By contrast, the retrieval module solves the dual task, i.e., retrieving relevant sentences expressing a specific relation, and hence it defines the probability $q_\phi(\mathbf{x}|y)$ for a mention-label pair $(\mathbf{x},y)$. In practice, since $q_\phi(\mathbf{x}|y) \propto q_\phi(\mathbf{x},y)$ when given a specific relation $y$, the retrieval module will focus on estimating $q_\phi(\mathbf{x},y)$ instead. The overall objective function is given below:
\begin{equation}
\label{eqn::obj}
    \mathbf{O}=\mathbf{O}_{P} + \mathbf{O}_{R} + \mathbf{O}_{U}
\end{equation}

There are three terms in total. In $\mathbf{O}_{P}$ and $\mathbf{O}_{R}$, we use the given labeled sentences to train the prediction module and the retrieval module, respectively. In the unsupervised part $\mathbf{O}_{U}$, we promote the collaboration of the two modules to annotate unlabeled sentences as additional training data. In the rest of this section, we will introduce the details of each key component in DualRE.

\subsection{Relation Prediction Module}

\begin{figure}[!t]
	\centering
	\includegraphics[width=0.47\textwidth]{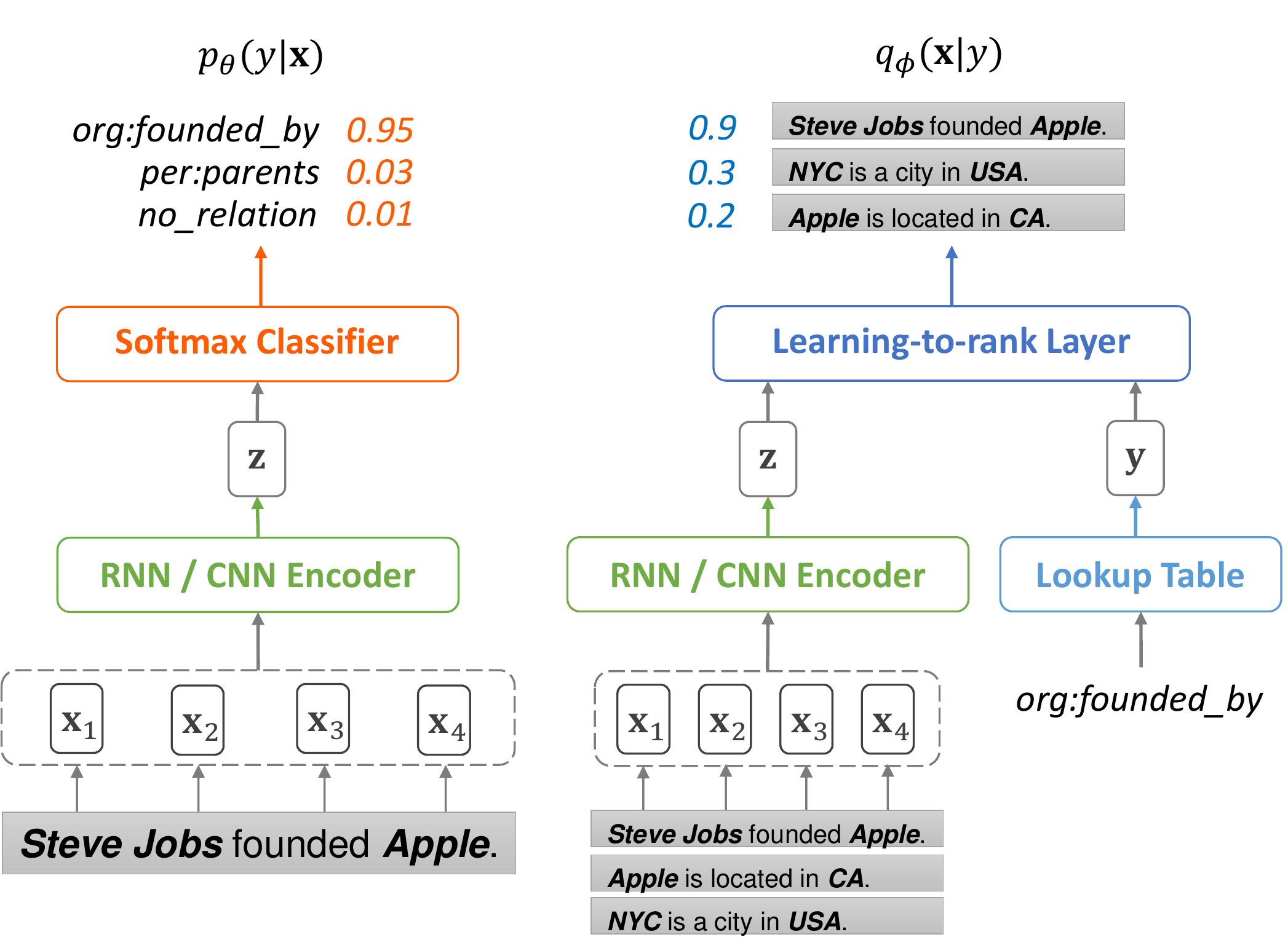}
    \vspace{0.2cm}
	\caption{\textbf{Model Design of the Two Modules in DualRE.} \textbf{Left}: The prediction module encodes the relation mention and outputs a probability distribution over all relations. \textbf{Right}: The retrieval module ranks the relation mentions given a query by predicting a relevance score $q_\phi(\mathbf{x}, y)$. Retrieving instances given a relation is essentially taking the top instances from the ranking list. Note that this figure does not illustrate how two modules mutually enhance each other using unlabeled data.}
	\label{fig::dualre_structure}
	\vspace{-0.7cm}
\end{figure}

In DualRE framework, the prediction module $\mathcal{P}_\theta$ focuses on the primal task, that is, predicting the relation expressed in a relation mention, as shown in Fig.~\ref{fig::dualre_structure}. We build the prediction module by first leveraging an encoder to encode the relation mention, and then adopting a softmax classifier to perform relation classification.

\smallskip
\noindent \textbf{Relation Mention Encoder.}
\label{sec::relation_mention_encoder}
The goal of relation mention encoders is to encode relation mentions into latent vectors, which can serve as features for prediction. A variety of encoders have been proposed for relation mention encoding, such as Long-Short Term Memory (LSTM) encoders~\cite{hochreiter1997long}, Piecewise Convolutional Neural Network (PCNN) encoders~\cite{zeng2015distant}, and Position-aware Recurrent Neural Network (PRNN) encoders~\cite{zhang2017position}. Among them, PRNN outperforms all other encoders on relation extraction, and therefore we use PRNN as our default encoder, but our framework can also adopt other encoders.


Formally, given a relation mention, PRNN considers the POS tagging, NER tagging, the position of each token (with regard to object and subject entities) as features. Then an LSTM cell is leveraged to learn the hidden representation of each word. Finally, a self-attention mechanism is utilized to 
weighted aggregate different hidden representations as the final encoding of the relation mention.

\smallskip
\noindent \textbf{Objective Function.}
\label{sec::prediction_module}
After encoding a relation mention, we feed it into a softmax classifier to predict the relation label of the mention. The prediction module can be trained using the given labeled data, in which we leverage the cross-entropy objective function, which is given below:
\begin{equation}
    \label{eqn::obj-p}
    \mathbf{O}_{P}=\mathbb{E}_{(\mathbf{x}, y) \in L}[\log p_\theta(y|\mathbf{x})]
\end{equation}

Given a relation mention $\mathbf{x}$ and its relation label $y$, the prediction module seeks to maximize the probability of $y$ for the mention $\mathbf{x}$.




\subsection{Sentence Retrieval Module}
\label{sec::retrieval_module}

The intended behavior of retrieval module $\mathcal{Q}_\phi$ is to select a set of relevant mentions given specific relations. In information retrieval domain, this is similar to the problem of retrieving a set of documents given a query. In our case, ``document'' and ``query'' correspond to relation mention and relation, respectively.

Learning-to-rank \cite{liu2009learning} is a type of model that aims to rank the documents according to its relevance to a query, thus we can select the top-ranking ones as retrieved documents. Learning-to-rank formulates such problems using joint probability $q_\phi(\mathbf{x}, y)$, which is in proportion to $q_\phi(\mathbf{x}|y)$ when the probability of observing query $q(y)$ is fixed. Similar to the prediction module, we can train the retrieval module by leveraging the labeled data and optimizing the following objective function:
\begin{equation}
    \label{eqn::obj-r-general}
    \mathbf{O}_{R}=\mathbb{E}_{(\mathbf{x}, y) \in L}[\log q_\phi(\mathbf{x},y)]
\end{equation}

However, directly optimizing $q_\phi(\mathbf{x}, y)$ is intractable, since we have to ensure $\sum_{(\mathbf{x}, y)} q_\phi(\mathbf{x}, y)=1$, which requires traversing over all possible pairs of $(\mathbf{x}, y)$ to calculate the partition function. Therefore, following existing learning-to-rank models, we seek not to enforce $q_\phi(\mathbf{x}, y)$ as a valid distribution. Instead, we let $q_\phi(\mathbf{x}, y)$ output a score to estimate the degree of \textit{relevance} of the pair $(\mathbf{x}, y)$. In literature, there are mainly two types of approaches to train such retrieval models, i.e., the pointwise approach~\cite{crammer2002pranking, li2008mcrank} and the pairwise approach~\cite{burges2005learning, cao2006adapting}, and we tried both in our experiment. Next, we introduce these approaches respectively. 


\smallskip \noindent \textbf{Pointwise Approach.}
Pointwise learning-to-rank models assume that each query-document pair is assigned a score that reflects their degree of relevance, which can be used to determine the rank for each document given a query.

In the task of retrieving relation mentions, for each relation mention $\mathbf{x}$, we have a label $y$ to indicate which relation it belongs to. Alternatively, we can regard such supervision signal as a binary score (0 or 1) that indicates whether a relation mention $\mathbf{x}$ is relevant to a specific relation $y \in \mathcal{R}$. In other words, each relation mention in labeled dataset is paired up with all relations, with a binary label to indicate if they are relevant.

The model design for the learning-to-rank retrieval module is illustrated in Fig. \ref{fig::dualre_structure}. Given a relation $y$ and a set of relation mentions $\{\mathbf{x}\}$ to be retrieved, we first encode each relation mention $\mathbf{x}$ as a vector $\mathbf{z}$. Each relation $y$ is mapped to its relation embedding $\mathbf{y}$ according to a relation embedding matrix $R$, which has the same dimension as $\mathbf{z}$. In learning-to-rank layer, pointwise approach takes the inner product $q_\phi(\mathbf{x}|y) = \sigma(\mathbf{z}^\top \mathbf{y})$ as the relevance score, where $\sigma(\cdot)$ denotes the sigmoid activation function.

\begin{equation}
\label{eqn::obj-pointwise}
    \mathbb{E}_{(\mathbf{x},y) \in L} [ \log \sigma(\mathbf{z}^\top \mathbf{y})] + \mathbb{E}_{(\mathbf{x},y') \notin L} [\log (1-\sigma(\mathbf{z}^\top \mathbf{y'})) ],
\end{equation}
where $(\mathbf{x},y)$ is a pair in the labeled dataset $L$, $(\mathbf{x},y')$ pairs up the relation mention $\mathbf{x}$ with incorrect relation $y'$, $\mathbf{z}$ is the mention encoding for $\mathbf{x}$, $\mathbf{y}$ and $\mathbf{y'}$ are embeddings of relation $y$ and $y'$.



\smallskip \noindent \textbf{Pairwise Approach.}
Pairwise learning-to-rank models assume that a relative order can be determined between the retrieved documents. For a set of documents, we can compare any pair of them and predict their relative order. We expect the model predictions (i.e. the set of relative orders) are consistent (no conflicting results such as $a > b, b > c$ but $c < a$), and thus the absolute order of each document in the ranked list can be derived. 

The core part of pairwise learning-to-rank models is the design of objective function that can characterize the relative order of paired documents. We adopt the approach in RankNet \cite{burges2005learning}, which proposes an objective function to minimize the number of inversions in ranking. An inversion happens when we rank a lower rated document above a higher rated one in a ranked list.


In the task of retrieving relation mentions, given a relation, we aim to rank all relevant mentions higher than the irrelevant ones. Similar to the design of pointwise learning-to-rank, we also encode each relation mention $\mathbf{x}$ and relation $y$ into vector representation $\mathbf{z}$ and $\mathbf{y}$ respectively. The only difference is how the objective function in learning-to-rank layer (as shown in Fig. \ref{fig::dualre_structure}) is calculated. We approximate $O_R$ in Eq.~\ref{eqn::obj-r-general} using the following term:
\begin{equation}
\label{eqn::retrieval-obj}
    \mathbb{E}_{(\mathbf{x}, y), (\mathbf{x}', y') \in L \times L}[ r(\mathbf{x}, \mathbf{x}', y) \log{\sigma(\mathbf{z}^T\mathbf{y} - \mathbf{z}'^T\mathbf{y})} ],
\end{equation}
where the order score $r(\mathbf{x}, \mathbf{x}', y)$ expresses the relative order of two relation mentions, given the relation $y$. Consider $\mathbf{x}$ as a positive instance for relation $y$, the order score will return 1 if $\mathbf{x}'$ is a negative instance. During optimization process, this indicates that we attribute $\mathbf{z}^Ty$ a higher score than $\mathbf{z}'^Ty$. If $\mathbf{x}'$ happens to be another positive instance, we cannot make assertion about their relative order, then the order score returns 1/2. During optimization process, this indicates that we assume the score $\mathbf{z}^Ty$ and $\mathbf{z}'^Ty$ are close.

\subsection{Interaction Between the Two Modules}
\label{sec::module_interaction}

So far, we only introduced the formulation of prediction and retrieval modules and how they can be trained in a supervised way (by maximizing $O_P$ and $O_R$). In semi-supervised setting, we try to incorporate unlabeled data, which leads to the introduction of $O_U$ in overall objective function. In DualRE, the formulation of this unsupervised objective function involves both prediction and retrieval modules. Intuitively, we hope to leverage the unlabeled data under the collaboration between both modules.

Formally, in $\mathbf{O}_{U}$, we seek to maximize the log-likelihood function of the unlabeled relation mentions:
\begin{equation}
\begin{split}
\label{eqn::obj-unsup}
    \mathbf{O}_{U}&=\mathbb{E}_{\mathbf{x} \in U}[\log p(\mathbf{x})] \\
                  &\geq \mathbb{E}_{\mathbf{x} \in U, y \sim p_\theta(y|\mathbf{x})}[\log \frac{q_\phi(\mathbf{x},y)}{p_\theta(y|\mathbf{x})}]
\end{split}
\end{equation}

The above inequality is derived from the EM algorithm~\cite{dempster1977maximum}. The two sides are equal when $p_\theta(y|\mathbf{x})=q_\phi(y|\mathbf{x})$. By maximizing the lower bound of $O_U$, we incorporate both modules in the optimization process. As we will see in Sec. \ref{sec::joint_learning}, we approach the optimization process by sampling from unlabeled data and accepting the instances that both modules agree on. These newly retrieved instances are regarded as pseudo-labeled data and enhance the training of both modules. Such process effectively leverages the selection quality of retrieved instances and boosts the overall model performance.


\section{Joint Learning of Prediction and Retrieval Modules}
\label{sec::joint_learning}

In Sec. \ref{sec::dualre_framework}, we introduce the formulation for three components ($O_P$, $O_R$ and $O_U$) of the overall objective function. This section will present the details on how we approach the optimization of the overall objective function $O$, and the learning algorithm.

\subsection{The EM-based Joint Optimization}
\label{sec::em-algo}
To optimize the objective function $O$, we follow the EM algorithm to alternatively optimize the prediction and retrieval modules. 

\smallskip
\noindent \textbf{E Step}. In this step, we update the prediction module $\mathcal{P}_\theta$ by fixing the retrieval module $\mathcal{Q}_\phi$. This corresponds to minimization of KL divergence $KL(p_\theta(y|\mathbf{x})||q_\phi(y|\mathbf{x}))$. Directly minimizing this objective function is difficult, thus we follow the wake-sleep algorithm~\cite{hinton1995wake} to minimize the reversed KL divergence $KL(q_\phi(y|\mathbf{x})||p_\theta(y|\mathbf{x}))$, which yields the same optimal solution, i.e., $p_\theta(y|\mathbf{x})=q_\phi(y|\mathbf{x})$. By integrating Eq.~\ref{eqn::obj-p} and Eq.~\ref{eqn::obj-unsup} as the overall objective function, and taking derivative with respect to $\theta$, the gradient for $\theta$ is calculated as:
\begin{equation}
\begin{split}
\label{eqn::grad-pp}
    \nabla_\theta \mathbf{O} &= \mathbb{E}_{(\mathbf{x}, y) \in L}[\nabla_\theta \log p_\theta(y|\mathbf{x})] \\
    &+ \mathbb{E}_{\mathbf{x} \in U, y \sim q_\phi(y|\mathbf{x})}[\nabla_\theta \log p_\theta(y|\mathbf{x})],
\end{split}
\end{equation}
where the gradient is calculated from both the labeled data and additional data annotated by the retrieval module $\mathcal{Q}_\phi$. However, only leveraging the retrieval module for annotation can be biased, so we may expect both modules to mutually correct each other. Towards this goal, we notice that $\mathbb{E}_{p_\theta(y|\mathbf{x})}[\nabla_\theta \log p_\theta(y|\mathbf{x})]=0$~\footnote{$\mathbb{E}_{p_\theta(y|\mathbf{x})}[\nabla_\theta \log p_\theta(y|\mathbf{x})]=\mathbb{E}_{p_\theta(y|\mathbf{x})}[\frac{\nabla_\theta p_\theta(y|\mathbf{x})}{p_\theta(y|\mathbf{x})}]=\sum \nabla_\theta p_\theta(y|\mathbf{x})=\nabla_\theta \sum p_\theta(y|\mathbf{x})=\nabla_\theta 1=0$}, and therefore the gradient can be modified as:
\begin{equation}
\begin{split}
\label{eqn::grad-p}
    \nabla_\theta \mathbf{O} &= \mathbb{E}_{(\mathbf{x}, y) \in L}[\nabla_\theta \log p_\theta(y|\mathbf{x})]\\
    &+ \mathbb{E}_{\mathbf{x} \in U, y \sim (q_\phi(y|\mathbf{x}) + p_\theta(y|\mathbf{x}))}[\nabla_\theta \log p_\theta(y|\mathbf{x})]
\end{split}
\end{equation}

Although the gradient remains the same as in Eq.~\ref{eqn::grad-pp}, the new gradient enables us to leverage both modules to collect additional labeled sentences, which usually has lower noise, and therefore leads to better performance.

\medskip 
\noindent \textbf{M Step}. In this step, we update the retrieval module $\mathcal{Q}_\phi$ by fixing the prediction module $\mathcal{P}_\theta$. The gradient with respect to $\phi$ can be similarly calculated as:
\begin{equation}
\begin{split}
\label{eqn::grad-q}
    \nabla_\phi \mathbf{O} &= \mathbb{E}_{(\mathbf{x}, y) \in L}[\nabla_\phi \log q_\phi(\mathbf{x},y)]\\
    &+ \mathbb{E}_{(\mathbf{x},y) \sim (p_\theta(\mathbf{x}, y) + q_\phi(\mathbf{x},y))}[\nabla_\phi \log q_\phi(\mathbf{x},y)]
\end{split}
\end{equation}

Again, directly calculating $\log q_\phi(\mathbf{x},y)$ is infeasible, and therefore in practice we use the pointwise term (Eq.~\ref{eqn::obj-pointwise}) or the pairwise objective function (Eq.~\ref{eqn::retrieval-obj}) to replace $\log q_\phi(\mathbf{x},y)$.

To summarize, the model alternates between optimizing the objective function with regard to $\phi$ and $\theta$, which is referred to as E and M step. During each step, we incorporate both labeled and unlabeled data.
Both E and M step require sampling annotated instances from unlabeled data. Noticing that samples from distribution $\mathbf{x} \in U, y \sim (q_\phi(y|\mathbf{x}) + p_\theta(y|\mathbf{x}))$ is equivalent to sampling from distribution $(\mathbf{x},y) \sim (p_\theta(\mathbf{x}, y) + q_\phi(\mathbf{x},y))$, we can keep a single version of sampled instances and apply to both steps. 



Next, we introduce an empirical sampling method that effectively generates high-quality training relation mentions (as additional supervision) to improve the performance.

\subsection{Instance Selection}
\label{sec::instance_selection}
    
Empirically, sampling from distribution $p_\theta(\mathbf{x},y) + q_\phi(\mathbf{x},y)$ can be approximated by taking the intersection set of predicted results from both modules.

In prediction module $\mathcal{P}_\theta$, we first select new instances from distribution $p_\theta(\mathbf{x}, y)$. If we assume a uniform distribution for relation mentions $p(\mathbf{x})$, we are essentially sampling from distribution $p_\theta(y|\mathbf{x}) \propto p_\theta(\mathbf{x}, y)$. Given the unlabeled instance $\mathbf{x}$ and its predicted relation $y$, we rank these instances according to the probability of predicting $y$ (i.e., $p(y|\mathbf{x})$) and take top-$k$ instances from the ranked list. This is similar to what we do in self-training models.

In retrieval module $\mathcal{Q}_\phi$, we sample $k$ instances from distribution $q_\phi(\mathbf{x}, y) = q_\phi(\mathbf{x}|y)q(y)$. The first term $q_\phi(\mathbf{x}|y)$ aligns with the intended behavior for retrieval module: given a relation $y$, retrieval module traverses through each unlabeled instance $\mathbf{x}$ and gives out a relevance score $q_\phi(\mathbf{x}, y) \propto q_\phi(\mathbf{x}|y)$. Therefore we can construct a ranked list for each relation $y$, in which top-$k_y$ instances are retrieved for this relation. We aggregate the instances from each relation to construct the retrieved set. The second term $q(y)$ can be regarded as prior knowledge that constrains the relation distribution for retrieved instances. Given that we retrieved $k$ instances which follows the reference relation distribution $q(y)$, we can determine the number of instances retrieved for each relation $k_y = kq(y)$.

Both modules collect a set of training instances, and the intersection of them are regarded as the final set. Intuitively, this is similar to majority vote or ensemble that consider multiple models when doing prediction.

\subsection{Instance Promotion}
\label{sec:instance_promotion}

To further eliminate the influence of incorrect annotations in retrieved instances, we experiment with two strategies to incorporate these instances in training phase.


\smallskip
\noindent \textbf{Equal promotion.}
Equal promotion treats the retrieved instances and original training data equally when calculating the loss function. This is the typical strategy for self-training models. The objective functions derived in Eq. \ref{eqn::grad-p} and \ref{eqn::grad-q} reflect this strategy.

\smallskip
\noindent \textbf{Weighted promotion.}
In this strategy, each instance is assigned a weight $\pi$ reflecting the confidence of its annotation, which is intended to be low when it's wrongly labeled. The weights are incorporated in training phase when promoting different instances. Considering Eq. \ref{eqn::grad-pp} when training the prediction module, the first term is intended for promoting instances in labeled set, which is assigned a weight of 1 and the formula remains unchanged. The second term is modified as follows:
\begin{equation}
    \mathbb{E}_{(\mathbf{x},y) \sim (p_\theta(\mathbf{x}, y) + q_\phi(\mathbf{x},y))}[\pi^p_\mathbf{x}\nabla_\phi \log q_\phi(\mathbf{x},y)],
\end{equation}
where $\pi^p_\mathbf{x}$ is the weight for pseudo-labeled instance (\textbf{x}, y). Similarly, we introduce the weight $\pi^q_\mathbf{x}$ in objective function $O_R$ for retrieval module. In our experiment, we parameterize the weights as $\pi^p_\mathbf{x} = p_\theta(y|\mathbf{x})^\alpha$ and $\pi^q_\mathbf{x} = q_\phi(\mathbf{x}, y)^\beta$, in which $\alpha$ and $\beta$ are hyperparmeters to tune in our model.



\subsection{The Joint Learning Algorithm}
\label{sec::joint_learning_algo}

We summarize the training algorithm for DualRE in Algo. \ref{algo::dualre}. The algorithm is basically an approximation of EM algorithm as described in Sec. \ref{sec::em-algo}. The whole process works in an iterative manner. In each iteration, we retrieve a new set of annotated instances using both prediction and retrieval modules. According to Sec. \ref{sec::instance_selection}, this is an empirical practice to sample from distribution $p_\theta(\mathbf{x}, y) + q_\phi(\mathbf{x},y)$. The retrieved instances, along with previous training data can be used for updating the prediction and retrieval module, which correspond to the E step and M step in EM algorithm. The iterative process stops when our model converges or the unlabeled data is exhausted.

\begin{algorithm}[t]
\SetAlgoLined
\KwIn{Labeled data $L = \{(\mathbf{x}_i, y_i)\}_{i=1}^{N_L}$, unlabeled data $U = \{\mathbf{x}_j\}_{j=1}^{N_U}$, the amount of data to retrieve each in each iteration $k$.}

Initialize: $L_U \leftarrow \emptyset$.\\
$P_\theta, Q_\phi \leftarrow$ Pretrain prediction and retrieval module using $L$.\\
\While{$U \neq \emptyset$ \textbf{and} not converge}{
    $L' \leftarrow$ Retrieve $k$ annotated instances $(\mathbf{x}, y)$ from $U$ (Sec. \ref{sec::instance_selection}).\\
    Remove instances $L'$ from $U$ and add them to $L_U$.\\
    \textrm{// Update prediction module:}\\
    Optimize $P_\theta$ using data from both $L$ and $L_U$ (Eq. \ref{eqn::grad-p}).\\
    \textrm{// Update retrieval module:}\\
    Optimize $Q_\phi$ using data from both $L$ and $L_U$ (Eq. \ref{eqn::grad-q}).\\
}
\caption{DualRE Learning Algorithm.}
\label{algo::dualre}
\end{algorithm}


%% file: 5-experiment.tex
\section{Experiments}
This section first introduces datasets and experimental settings, and then presents performance comparison results with baseline methods as well as case studies of various model components in order to validate different design choices in DualRE.

\subsection{Dataset}

\begin{table} [!tb]
	\begin{center}
	    \scalebox{0.83}{
		\begin{tabular}{C{1.8cm}R{1.1cm} R{1.0cm} R{1.0cm} R{1.6cm} R{1.4cm}}\cline{1-6}
		    \toprule
		    \textbf{Dataset} & \textbf{\# Train} & \textbf{\# Dev} & \textbf{\# Test} & \textbf{\# Relations} & \textbf{\% Neg.} \\
		    \midrule
		    SemEval~\cite{hendrickx2009semeval} & 7,199 & 800 & 1,864 & 19 & 17.40 \\
		    TACRED~\cite{zhang2017position} & 75,049 & 25,763 & 18,659 & 42 & 78.68 \\
		    \bottomrule
	    \end{tabular}
        }
	\end{center}
	\caption{\textbf{Statistics for SemEval and TACRED dataset.} We present the number of relation mentions in the train/dev/test sets. In particular, \% Neg. implies the percentage of ``negative" relation mentions (i.e., relation mentions that are labeled as ``\textsf{no\_relation}'').}
    \label{tab::data}
\end{table}

\begin{table*}[!t]
\begin{center}
\scalebox{0.73}{
\begin{tabular}{L{0.25\linewidth}C{0.1\linewidth}C{0.1\linewidth}C{0.1\linewidth}C{0.1\linewidth}C{0.1\linewidth}C{0.1\linewidth}C{0.1\linewidth}C{0.1\linewidth}C{0.1\linewidth}}
    \toprule
    \textbf{Methods / \% Labeled Data} & \multicolumn{3}{c}{\textbf{5\%}} & \multicolumn{3}{c}{\textbf{10\%}} & \multicolumn{3}{c}{\textbf{30\%}} \\
     & Precision & Recall & $F_1$ & Precision & Recall & $F_1$ & Precision & Recall & $F_1$ \\
    \midrule
    LSTM \cite{hochreiter1997long} & 25.31 $\pm$ 2.16 & 20.72 $\pm$ 3.91 & 22.71 $\pm$ 3.31 & 36.91 $\pm$ 6.10 & 29.85 $\pm$ 7.25 & 32.92 $\pm$ 6.71 & 64.83 $\pm$ 0.69 & 63.03 $\pm$ 0.67 & 63.91 $\pm$ 0.66 \\
    PCNN \cite{zeng2015distant} & 42.95 $\pm$ 4.69 & 40.79 $\pm$ 4.59 & 41.84 $\pm$ 4.63 & 53.78 $\pm$ 1.51 & 49.11 $\pm$ 2.22 & 51.32 $\pm$ 1.74 & 64.54 $\pm$ 0.58 & 62.98 $\pm$ 0.48 & 63.75 $\pm$ 0.33 \\
    PRNN \cite{zhang2017position} & 56.16 $\pm$ 1.32 & 54.87 $\pm$ 1.49 & 55.49 $\pm$ 0.90 & 61.70 $\pm$ 1.16 & 63.61 $\pm$ 2.07 & 62.63 $\pm$ 1.42 & 69.66 $\pm$ 2.19 & 68.76 $\pm$ 2.60 & 69.14 $\pm$ 1.02 \\
    \hline
    \hline
    Mean-Teacher (PRNN) \cite{tarvainen2017mean}  & 53.71 $\pm$ 4.43 & 49.54 $\pm$ 3.29 & 51.51 $\pm$ 3.58 & 62.43 $\pm$ 1.28 & 60.34 $\pm$ 0.62 & 61.36 $\pm$ 0.75 & 68.65 $\pm$ 0.64 & 69.84 $\pm$ 0.65 & 69.24 $\pm$ 0.56\\
    Self-Training (PRNN) \cite{rosenberg2005semi} & 56.47 $\pm$ 1.11 & 56.14 $\pm$ 1.33 & 56.30 $\pm$ 0.96 & 64.27 $\pm$ 2.37 & 63.48 $\pm$ 2.02 & 63.79 $\pm$ 0.28 & 68.95 $\pm$ 0.68 & 72.63 $\pm$ 0.82 & 70.74 $\pm$ 0.58 \\
    RE-Ensemble (PRNN) & 58.77 $\pm$ 0.58 & 58.50 $\pm$ 0.97 & 58.63 $\pm$ 0.62 & 65.10 $\pm$ 0.84 & 64.57 $\pm$ 0.54 & 64.83 $\pm$ 0.61 & 70.26 $\pm$ 0.92 & 73.20 $\pm$ 1.22 & 71.69 $\pm$ 0.47 \\
    \midrule
    DualRE-Pairwise (PRNN) & 59.76 $\pm$ 0.47 & 63.36 $\pm$ 0.77 & \textbf{61.51 $\pm$ 0.56} & 64.39 $\pm$ 0.75 & 67.70 $\pm$ 0.80 & 66.00 $\pm$ 0.48 & 70.05 $\pm$ 0.53 & 74.83 $\pm$ 0.88 & 72.36 $\pm$ 0.60 \\
    DualRE-Pointwise (PRNN) & 58.73 $\pm$ 1.50 & 62.23 $\pm$ 1.93 & 60.43 $\pm$ 1.67 & 64.50 $\pm$ 1.14 & 67.67 $\pm$ 1.66 & \textbf{66.03 $\pm$ 1.00} & 70.03 $\pm$ 0.74 & 74.87 $\pm$ 0.75 & \textbf{72.36 $\pm$ 0.35}\\
    \hline
    \hline
    RE-Gold (PRNN w. gold labels)\raise0.3ex\hbox{*} & 72.57 $\pm$ 1.47 & 74.65 $\pm$ 1.98 & 73.56 $\pm$ 0.31 & 71.40 $\pm$ 1.42 & 76.72 $\pm$ 0.64 & 73.95 $\pm$ 0.50 & 72.98 $\pm$ 0.96 & 78.86 $\pm$ 0.76 & 75.80 $\pm$ 0.24 \\
    \bottomrule
\end{tabular}
}
\end{center}
\vspace{-0.2cm}
\caption{\textbf{Performance comparison on SemEval~\cite{hendrickx2009semeval} with various amounts of labeled data and 50\% unlabeled data}. We report the mean and standard deviation of the evaluation metrics by conducting 5 runs of training and testing using different random seeds. DualRE outperforms all the baseline methods.}
\label{tab::semeval_results}
\end{table*}

We choose two public relation extraction datasets with different characteristics to validate the effectiveness of our proposed method.
(1) \textbf{SemEval} \cite{hendrickx2009semeval}. SemEval 2010 Task 8 provides a standard benchmark dataset which is widely adopted for evaluating relation extraction systems. It has around 10,000 relation mentions in total (where entity mentions are given), with 19 relations (including ``\textsf{no\_relation}'').
(2) \textbf{TACRED} \cite{zhang2017position}. The TAC Relation Extraction Dataset is a large-scale crowd-sourced relation extraction dataset following the TAC KBP relation schema\footnote{\small \url{https://tac.nist.gov/2017/KBP}}. The corpora are collected from all the prior TAC KBP shared tasks. It has more than 100,000 relation mentions with relations categorized into 42 classes (including ``\textsf{no\_relation}'').

Table~\ref{tab::data} shows the overall statistics for these two datasets, which are significantly different in terms of dataset size, number of relations, and percentage of negative instances. As one can observe from the table, TACRED dataset is far more complicated than SemEval, since it has more types of relations and more skewed distribution between positive and negative instances. We conduct experiments on both datasets so as to validate the effectiveness of our DualRE model learned with datasets of different characteristics.

\subsection{Compared Methods}
As the proposed DualRE framework is general and can be integrated with various supervised ``base" model of relation extraction, we first compare several widely used supervised models, then adopt the best performing one as the base model and compare DualRE with other representative semi-supervised methods. All models mentioned below are re-implemented and integrated into a unified interface for efficient experiment conducting.

For base models proposed for supervised relation extraction, we select the following models and train them only on the labeled dataset (i.e., set $L$):
(1) \textbf{LSTM} \cite{hochreiter1997long}: The Long Short-term Memory (LSTM) network is a variant of basic Recurrent Neural Network (RNN). We can simply adopt LSTM without attention mechanism to encode the word sequence and classify it into predefined relation categories based on the sentence representation. 
(2) \textbf{PCNN} \cite{zeng2015distant}: Different from the encoding part of LSTM RE model, Piecewise Convolutional Neural Network (PCNN) uses CNN with piecewise max pooling to encode the word sequence. Position embeddings are used to provide the model with the positions of subject and object entities in the sentence. 
(3) \textbf{PRNN} \cite{zhang2017position}: Integrating position-aware attention mechanism, the Position-aware Recurrent Neural Network (PRNN) is proposed specifically for supervised relation extraction and has achieved state-of-the-art performance on TACRED dataset. Details of PRNN model are described in Sec.~\ref{sec::prediction_module}.

As PRNN has achieved state-of-the-art performance in prior work, we construct our DualRE model based on PRNN, and compare with other semi-supervised classification methods which also adopt PRNN as the base model for fair evaluation. For semi-supervised baselines, we select several representative methods from different categories of semi-supervised learning approaches (See Sec.~\ref{sec::related_work} for details), shown as follows.
(4) \textbf{Self-Training} \cite{rosenberg2005semi}: Self-Training is a semi-supervised method that uses a single model's predictions on unlabeled data to iteratively re-train and improve the model itself. In each iteration, the model is first used to label the unlabeled instances, then the most confident pseudo-labeled instances are added to the labeled set, which are further used to re-train the model in the next iteration. The iterative training procedure stops when the unlabeled data is exhausted.
(5) \textbf{Mean-Teacher} \cite{tarvainen2017mean}: Mean-Teacher belongs to the category of self-ensembling methods. It is first proposed for semi-supervised image classification task. The main idea is to encourage different variants of the model to make consistent predictions over similar inputs. A perturbation-based loss, and the original training loss, are jointly optimized to utilize the information provided by unlabeled data. In our implementation, the dropout strategy \cite{srivastava2014dropout,gal2016theoretically} is applied to generate the perturbed input.

In addition to the existing semi-supervised methods introduced above, we propose a variant of DualRE model as another baseline model, to further validate the effectiveness of adopting a dual retrieval model in contrast to ensembling two prediction models with identical configuration, i.e., both modules use PRNN as the same base model but with different model parameters.
(6) \textbf{RE-Ensemble}: By replacing the retrieval module in the proposed DualRE framework with the same prediction module (with a new set of model parameters), we design the RE-Ensemble model that selects the promoted instances based on the agreement of two prediction modules in each iteration. The structure and size of these two models are exactly the same, but they are independently initialized at random to make sure they are not identical.

For our proposed DualRE method, we focus on studying the performance of two model variants based on different learning-to-rank objectives in the retrieval module. The models are trained on both the labeled data and the unlabeled data, as detailed below:
(1) \textbf{DualRE-Pointwise}: A DualRE model that uses pointwise learning-to-rank loss~\cite{cossock2006subset} in the retrieval module.
(2) \textbf{DualRE-Pairwise}: A DualRE model that uses pairwise learning-to-rank loss~\cite{burges2005learning} in the retrieval module.

Finally, we present another model to demonstrate the upper bound of the performance for all the semi-supervised methods. This model is learned using both the labeled dataset and the ``unlabeled'' dataset.
\textbf{RE-Gold}: It trains the PRNN base model using both the labeled relation mentions and the gold labels of relation mentions from the unlabeled set. This represents the setting that labels over all unlabeled instances can be perfectly predicted, which gives an upper bound performance.

\begin{table*}[!t]
\begin{center}
\scalebox{0.73}{

\begin{tabular}{L{0.25\linewidth}C{0.1\linewidth}C{0.1\linewidth}C{0.1\linewidth}C{0.1\linewidth}C{0.1\linewidth}C{0.1\linewidth}C{0.1\linewidth}C{0.1\linewidth}C{0.1\linewidth}}
    \toprule
    \textbf{Methods / \% Labeled Data} & \multicolumn{3}{c}{\textbf{3\%}} & \multicolumn{3}{c}{\textbf{10\%}} & \multicolumn{3}{c}{\textbf{15\%}} \\
     & Precision & Recall & $F_1$ & Precision & Recall & $F_1$ & Precision & Recall & $F_1$ \\
    \midrule
    LSTM \cite{hochreiter1997long}  & 40.76 $\pm$ 6.62 & 23.05 $\pm$ 4.48 & 28.62 $\pm$ 3.01 & 50.56 $\pm$ 0.98 & 43.09 $\pm$ 1.26 & 46.51 $\pm$ 0.90 & 55.11 $\pm$ 0.56 & 44.41 $\pm$ 0.58 & 49.18 $\pm$ 0.52\\
    PCNN \cite{zeng2015distant}  & 58.16 $\pm$ 7.74 & 37.49 $\pm$ 6.16 & 44.64 $\pm$ 1.59 & 64.32 $\pm$ 7.78 & 42.06 $\pm$ 4.94 & 50.16 $\pm$ 1.15 & 67.10 $\pm$ 0.44 & 42.88 $\pm$ 0.42 & 52.32 $\pm$ 0.30 \\
    PRNN \cite{zhang2017position} & 48.93 $\pm$ 5.72 & 33.05 $\pm$ 2.19 & 39.16 $\pm$ 0.90 & 53.44 $\pm$ 2.82 & 51.77 $\pm$ 1.88 & 52.49 $\pm$ 0.64 & 58.88 $\pm$ 2.32 & 51.30 $\pm$ 2.04 & 54.76 $\pm$ 0.93\\
    \hline
    \hline
    Mean-Teacher (PRNN) \cite{tarvainen2017mean}  & 53.08 $\pm$ 3.55 & 41.81 $\pm$ 0.61 & \textbf{46.74 $\pm$ 1.70} & 58.53 $\pm$ 2.56 & 50.08 $\pm$ 1.14 & 53.94 $\pm$ 0.91 & 57.90 $\pm$ 1.09 & 52.64 $\pm$ 0.97 & 55.13 $\pm$ 0.05\\
    Self-Training (PRNN) \cite{rosenberg2005semi} & 49.89 $\pm$ 1.05 & 39.23 $\pm$ 2.26 & 43.86 $\pm$ 1.26 & 56.54 $\pm$ 0.72 & 53.00 $\pm$ 0.49 & 54.71 $\pm$ 0.09 & 60.09 $\pm$ 0.43 & 54.77 $\pm$ 0.55 & 57.31 $\pm$ 0.47\\
    RE-Ensemble (PRNN) & 56.48 $\pm$ 0.95 & 36.90 $\pm$ 0.91 & 44.62 $\pm$ 0.39 & 61.26 $\pm$ 0.58 & 52.51 $\pm$ 0.56 & 55.54 $\pm$ 0.29 & 60.76 $\pm$ 0.78 & 55.00 $\pm$ 1.04 & 57.72 $\pm$ 0.38 \\
    \midrule
    DualRE-Pairwise (PRNN) & 58.97 $\pm$ 0.96 & 34.55 $\pm$ 1.18 & 43.55 $\pm$ 0.67 & 63.10 $\pm$ 0.94 & 48.91 $\pm$ 0.93 & 55.09 $\pm$ 0.25 & 60.99 $\pm$ 1.39 & 54.04 $\pm$ 0.46 & 57.30 $\pm$ 0.81 \\
    DualRE-Pointwise (PRNN) & 52.76 $\pm$ 2.58 & 38.99 $\pm$ 2.08 & 44.73 $\pm$ 0.66 & 61.61 $\pm$ 1.30 & 52.30 $\pm$ 0.89 & \textbf{56.56 $\pm$ 0.42} & 60.66 $\pm$ 1.57 & 56.65 $\pm$ 0.37 & \textbf{58.58 $\pm$ 0.69}\\
    \hline
    \hline
    RE-Gold (PRNN w. gold labels)\raise0.3ex\hbox{*}  & 64.38 $\pm$ 0.46 & 60.35 $\pm$ 0.81 & 62.30 $\pm$ 0.29 & 65.88 $\pm$ 0.66 & 61.65 $\pm$ 0.42 & 63.70 $\pm$ 0.53 & 66.95 $\pm$ 2.78 & 59.97 $\pm$ 3.12 & 63.13 $\pm$ 0.56 \\
    \bottomrule
\end{tabular}
}
\end{center}
\vspace{-0.2cm}
\caption{\textbf{Performance comparison on TACRED~\cite{zhang2017position} with various amounts of labeled data and 50\% unlabeled data}. We report the mean and standard deviation of the evaluation metrics by conducting 3 runs of training and testing using different random seeds. DualRE outperforms all the baseline methods except Mean-Teacher at one data point.}
\label{tab::tacred_results}
\end{table*}

\begin{figure*}
\vspace{-0.2cm}
	\subfigure{\includegraphics[width=0.31\linewidth]{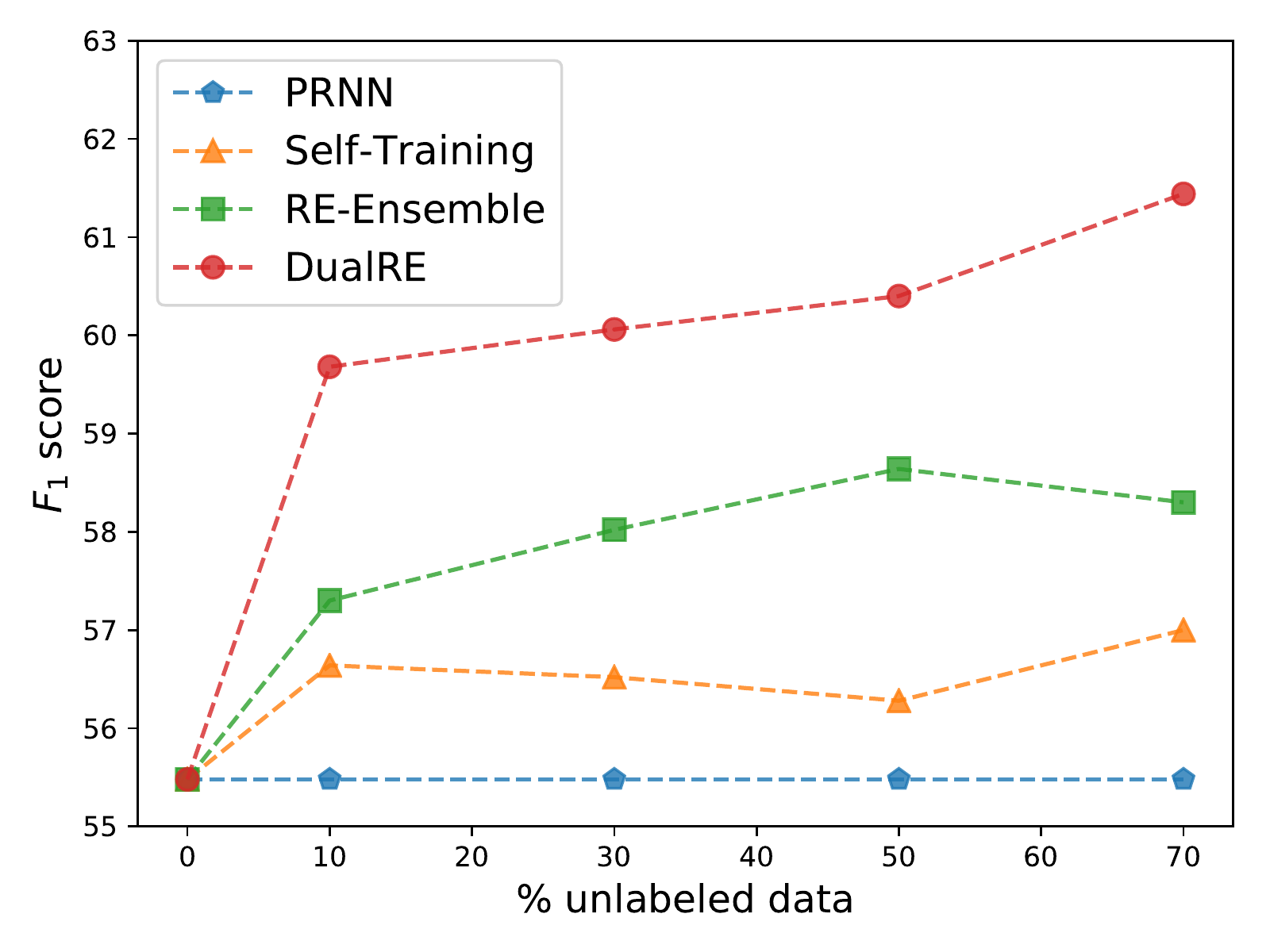}}
	\subfigure{\includegraphics[width=0.31\linewidth]{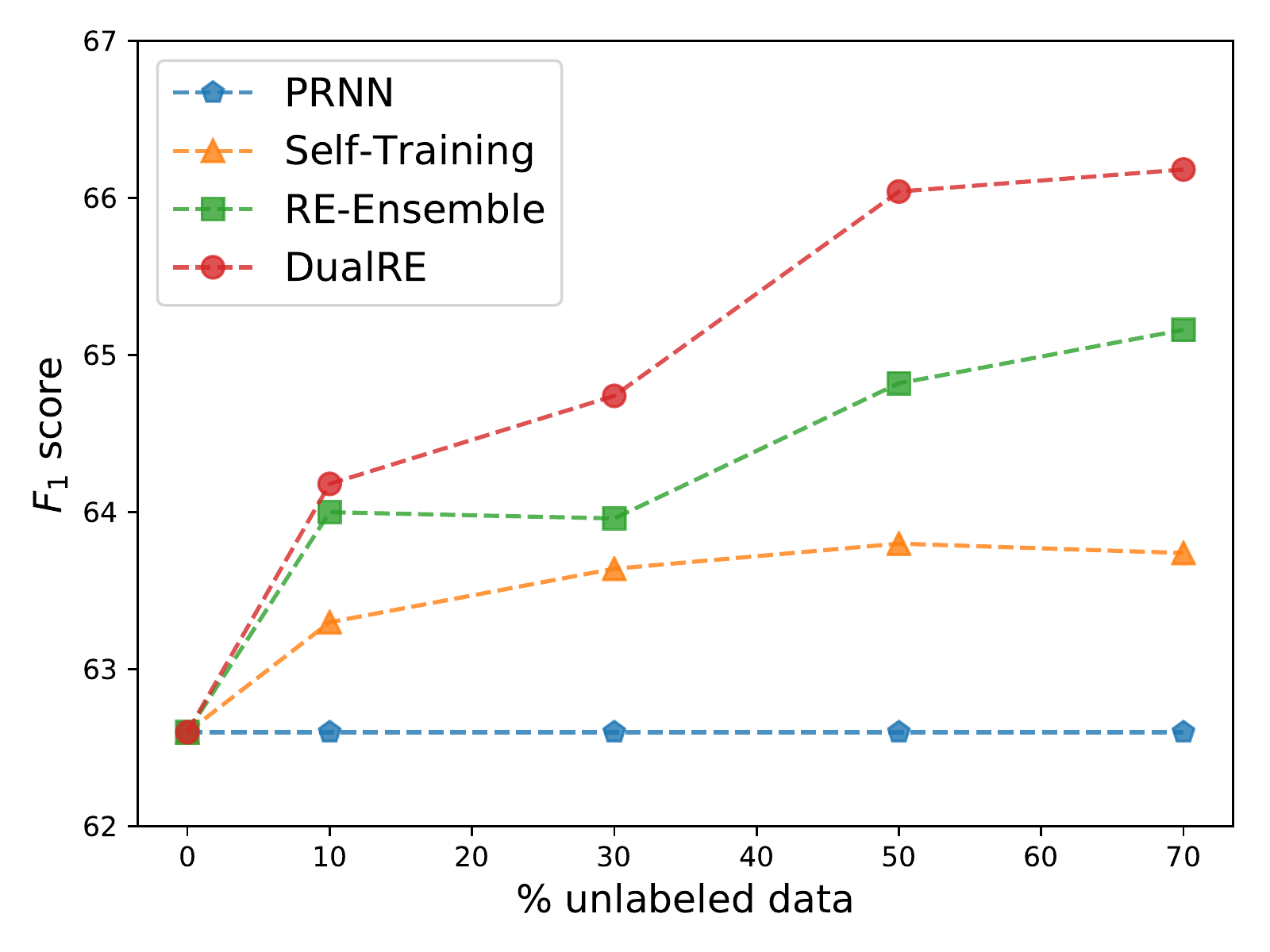}}
	\subfigure{\includegraphics[width=0.31\linewidth]{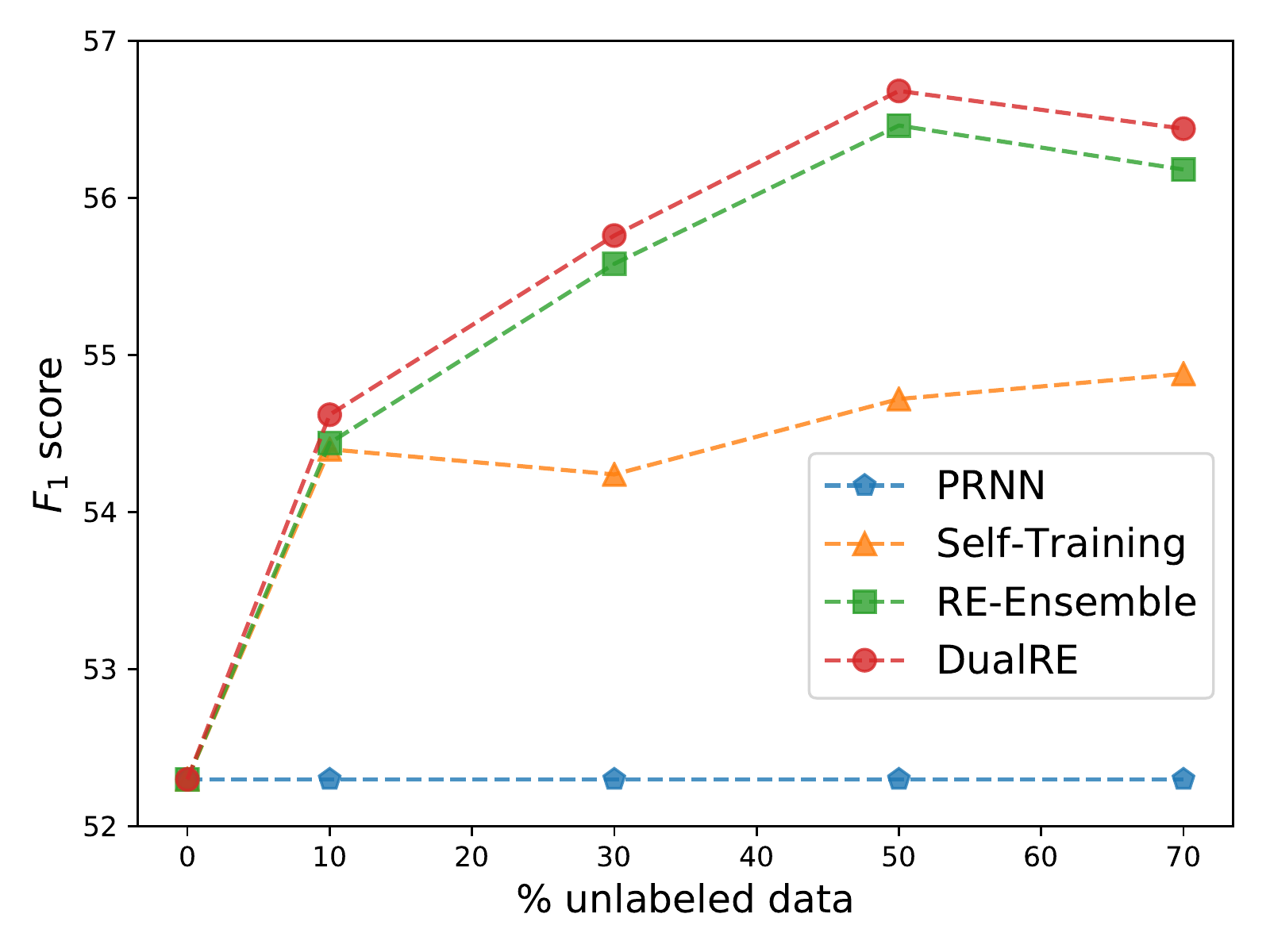}}
	\caption{\textbf{Performance w.r.t. the amount of unlabeled data.} \textbf{Left:} 5\% labeled data on SemEval; \textbf{Middle:} 10\% labeled data on SemEval; \textbf{Right:} 10\% labeled data on TACRED. In most cases, increasing unlabeled data results in better performance, but too much unlabeled data may hurt the performance.}
	\label{fig::performance_curve}
\end{figure*}

\vspace{-0.1cm}
\subsection{Experimental Settings}

\noindent \textbf{Data Preparation}. 
To study the performance of semi-supervised relation extraction models with various labeled and unlabeled dataset sizes, we sample from the original training set and construct various sizes of labeled and unlabeled set. More specifically, for SemEval dataset, we sample 5\%, 10\% and 30\% training data as labeled sets, the ratios amounts to 3\%, 10\% and 15\% in TACRED. In both datasets, we sample 50\% training data as unlabeled set, whose ground-truth label is unavailable to models. During the sample process, we perform stratified sampling to ensure the label distribution remains unchanged. Duplication in labeled and unlabeled set is avoided as well. Note that since SemEval does not have official development set, we sample 10\% of the training set as the development set, before the above data preparation process.


\smallskip
\noindent \textbf{Data Processing}. Since the SemEval dataset does not contain named entity (NE) tag for each relation mention, which is essential for our PRNN base model, we apply the Stanford CoreNLP toolkit~\cite{manning2014stanford} to extract the above information.

\smallskip 
\noindent \textbf{Hyperparameter Tuning}. For base models (LSTM, PCNN, PRNN), we follow the default hyperparameter settings used in previous works~\cite{zeng2015distant,zhang2017position}. For fair comparison, both DualRE and other baseline methods are built on top of the same base models. For Mean-Teacher, we perform grid search for weight of consistency loss (select from $\{0.1, 1, 10\}$) and number of labeled instances in a mini-batch with size 50 (select from $\{1, 3, 5\}$). For DualRE, we perform greedy search for retrieval module dropout (select from $\{0.1, 0.3, 0.5, 0.7, 0.9\}$), score weight $\alpha$, $\beta$ (select from $\{0.5, 1, 2\}$) and label distribution (select from top-$k$ to top-$7k$ distribution and true distribution). The best hyperparameters are chosen according to model performance on the development set.



For semi-supervised models that promote instances in an iterative manner (Self-Training, RE-Ensemble and DualRE), we fix the number of retrieved instances per iteration $k$ for fair comparison. Since DualRE model takes the joint selection of prediction and retrieval modules, retrieving a fixed amount of instances $k$ is not guaranteed. To solve this problem, we starts with a retrieval upper bound $k'=k$ for both modules to select instances. We iteratively increase the upper bound ($k' = 1.25k'$) until we can sample $k$ unique instances from the joint set. In our experiments, $k$ is set to 10\% of the number of original unlabeled instances, which means that the unlabeled data is exhausted after 10 iterations.


\smallskip \noindent \textbf{Evaluation Metric}.
Following previous work in relation extraction \cite{zhang2017position,lin2016neural}, we consider $F_1$ score as the main metric while precision and recall serve as auxiliary metrics. Note that the correct prediction for \textsf{no\_relation} is ignored during the calculation. 

\subsection{Experiments and Performance Study}


Table~\ref{tab::semeval_results} and \ref{tab::tacred_results} show the model performance on both SemEval and TACRED dataset when 50\% unlabeled data and various labeled data are available. We conduct several runs of training and testing under different random seeds and report the mean and standard deviation in the result.


\smallskip \noindent \textbf{Comparing DualRE with Different Methods and Its Variants}. Comparing the performance of different models in Table \ref{tab::semeval_results} and \ref{tab::tacred_results}, the first conclusion we draw is that PRNN outperforms other base models in almost all the data points, which demonstrates the effectiveness of the position-aware encoder with the attention mechanism. This justifies the reason for choosing PRNN as base model for all semi-supervised models.

Secondly, moving on to semi-supervised baseline models, we observe a consistent performance gain when comparing RE-Ensemble with Self-Training. This is a strong evidence that ensemble of single models can boost the performance. However, Mean-Teacher as another self-ensembling method, does not have consistent gain over other models: It shows a clear win in TACRED dataset with 3\% labeled data, but falls far behind in all other data points. The reason behind such phenomenon is still under investigation, but Mean-Teacher is clearly not as stable and strong as other semi-supervised baseline models.



Thirdly, our DualRE-Pointwise model outperforms all the baseline models except on TACRED with 3\% labeled data. Its overall performance shows the effectiveness of DualRE framework. From the results, we attribute the performance gain to two design choices: (1) Ensemble method. Both DualRE-Pointwise and RE-Ensemble apply such strategy. By adding another prediction module, RE-Ensemble can already outperform its counterpart (Self-Training). (2) Introduction of dual task. Instead of replicating another prediction module as in RE-Ensemble, DualRE-Pointwise introduces the dual retrieval module. Two modules are jointly trained and complement each other. Thus we observe a consistent performance gain of DualRE-Pointwise over RE-Ensemble.

Lastly, speaking of the variant of DualRE, DualRE-Pairwise has comparable or even better performance as DualRE-Pointwise on the SemEval dataset but not on TACRED. Since DualRE-Pointwise generally performs better than DualRE-Pairwise, we use ``DualRE'' to refer to DualRE-Pointwise model in the following sections.

\smallskip \noindent \textbf{Performance on Different Datasets}.
One can clearly observe that the advantage of DualRE-Pairwise model over other baseline models becomes smaller on the TACRED dataset, comparing to SemEval. Considering the complexity and imbalance distribution in TACRED dataset over SemEval, it is more difficult to learn a strong retrieval module, especially for those rare relations. Thus DualRE models are not significantly better than other models.




\begin{table} [!t]
	\begin{footnotesize}
	\begin{center}
		\begin{tabular}{L{3.8cm}C{1.7cm} C{1.7cm}}
		    \toprule
		    \textbf{Instance Weight} & \textbf{$F_1$ (Dev)} & \textbf{$F_1$ (Test)} \\
		    \midrule
		    $\alpha$=0, $\beta$=0 (Equal Promotion) & 62.40 $\pm$ 0.52 & 64.42 $\pm$ 0.82 \\
		    $\alpha$=1, $\beta$=1 & 63.84 $\pm$ 0.23 & 65.98 $\pm$ 0.46 \\
            $\alpha$=1, $\beta$=2 & 64.22 $\pm$ 0.26 & 66.02 $\pm$ 0.15 \\
		    $\alpha$=0.5, $\beta$=2 & \textbf{64.28 $\pm$ 0.46} & \textbf{66.04 $\pm$ 1.00} \\
		    \bottomrule
	    \end{tabular}
	\end{center}
	\end{footnotesize}
	\vspace{-0.2cm}
	\caption{\textbf{Comparison of Different Weighing Methods for Instance Promotion.} We test with 10\% labeled data and 50\% unlabeled data on the SemEval dataset. Weighted promotion leads to significant improvement as compared with equal promotion.}
	\label{tab::instance_promotion_exp}
	\vspace{-0.4cm}
\end{table}

\smallskip \noindent \textbf{Performance on Different Amounts of Labeled Data}.
From Table \ref{tab::semeval_results} and \ref{tab::tacred_results}, when we fix the amount of unlabeled data to 50\% of the original training set, we can observe that increasing the amount of labeled data effectively leverage the model performance. Therefore, adding more labeled data (if accessible) is an efficient approach to boost the performance. 

\smallskip \noindent \textbf{Performance on Different Amounts of Unlabeled Data}.
In specific domains where human-annotated training instances are expensive to obtain, we aim to leverage a large amount of unlabeled data to boost the performance of our model given limited amount of labeled data. Therefore, for semi-supervised learning, we may wonder whether more unlabeled data will always help. To investigate such a problem, we fix the amount of labeled data and compare the performance under different amounts of unlabeled data. More specifically, we investigate the model performance in SemEval and TACRED dataset, with 5\% or 10\% labeled data presented. In our experiment, 10\%, 30\%, 50\% and 70\% unlabeled data are available.



We summarize the results of DualRE and other methods in Fig. \ref{fig::performance_curve}. One can observe that most models benefit from larger amount of unlabeled data in the range of 10\% to 50\%. However, further increasing unlabeled data to 70\% doesn't result in further improvement for all models. Half of them perform even worse with more unlabeled data. Therefore, it is not necessarily beneficial for semi-supervised models to take in as much unlabeled data as possible. 


\subsection{Case Study}

\smallskip \noindent \textbf{Effectiveness of Weighted Promotion}.
We experiment with two different ways of assigning weights to pseudo-labeled instances while training DualRE model. Specifically, different combinations of $\alpha$ and $\beta$ are explored, according to Section \ref{sec:instance_promotion} . Assigning $\alpha$=0 and $\beta$=0, for instance, is to promote all instances equally in training. Results in Table \ref{tab::instance_promotion_exp} indicate that by using weighted promotion, we can get better results than equal promotion.

\begin{figure}
	\centering
    \subfigure{\includegraphics[width=0.49\linewidth]{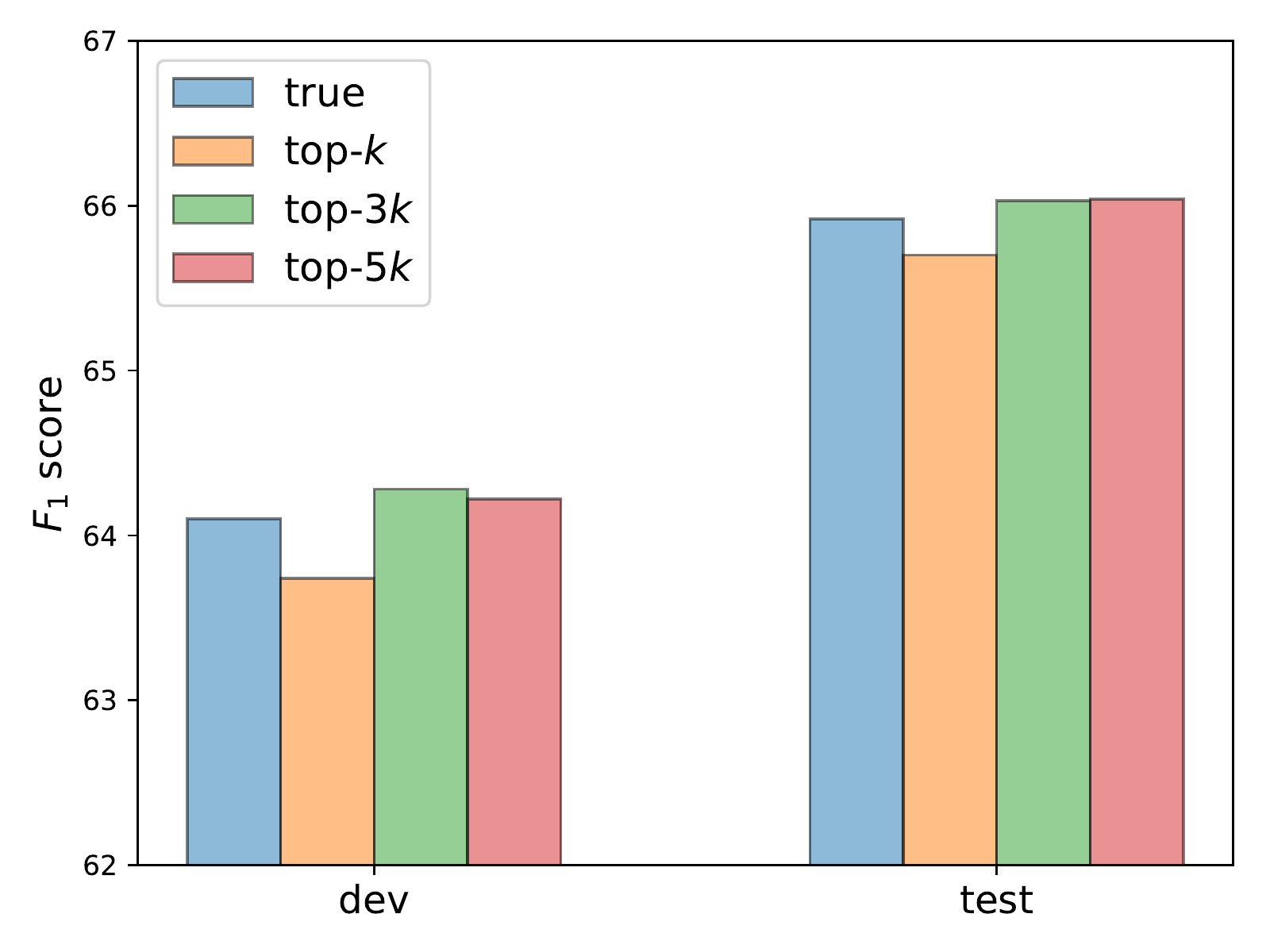}}
	\subfigure{\includegraphics[width=0.49\linewidth]{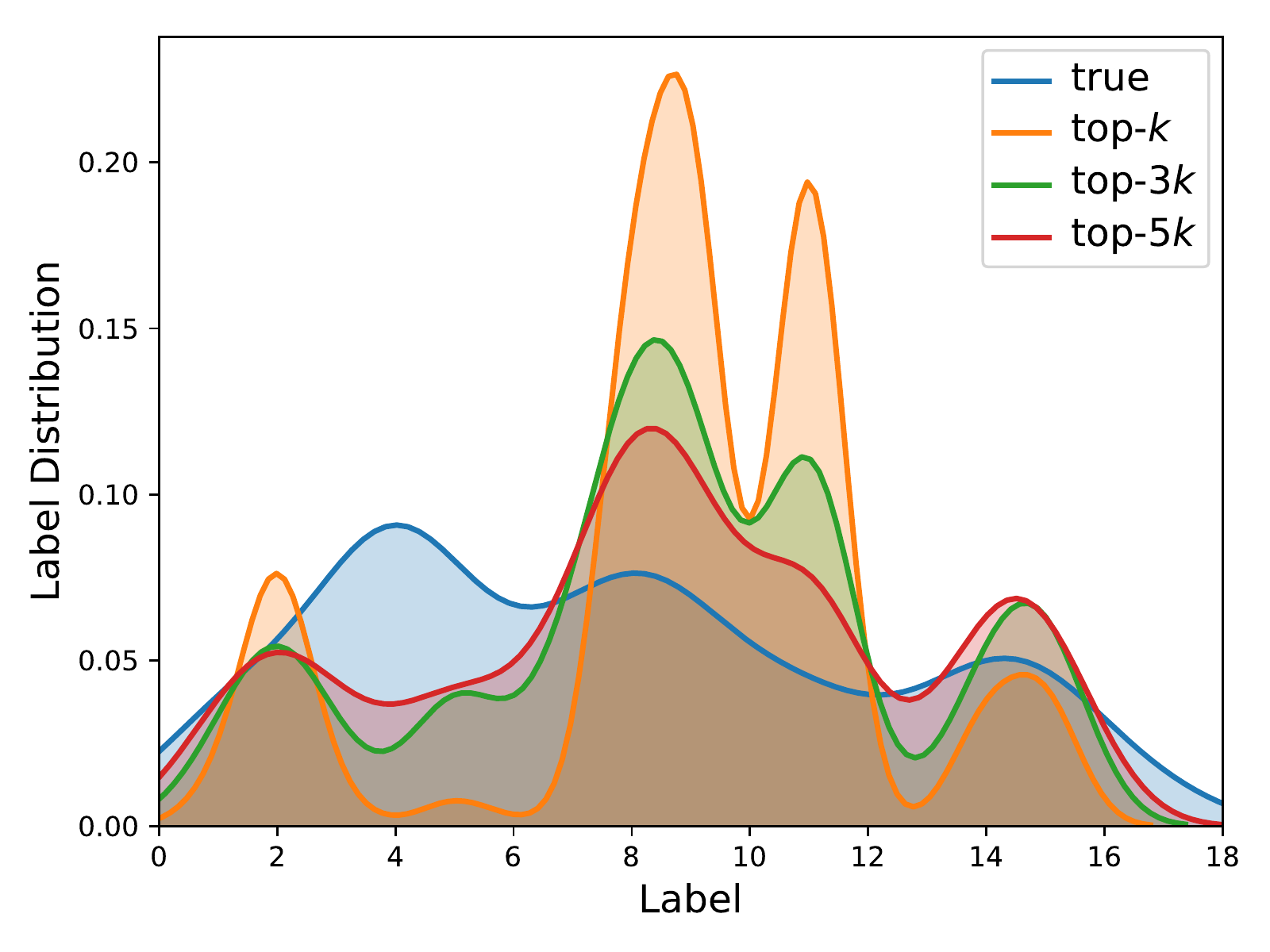}}
	\caption{\textbf{Case Study on Retrieval Module in DualRE.} \textbf{Left:} Comparison of using different label distributions as the retrieval query; \textbf{Right:} A visualization of different label distributions in the first iteration. Both are evaluated with 10\% labeled data and 50\% unlabeled data on the SemEval dataset.}
	\label{fig::label_distribution_results}
	\vspace{-0.5cm}
\end{figure}

\smallskip \noindent \textbf{Choice of Reference Relation Distribution}.
In DualRE, when selecting instances to promote, the retrieval module takes in a reference relation distribution $q(y)$ as described in Sec.~\ref{sec::instance_selection}. Practically, we assign $q(y)$ with label distribution from available dataset. Two options are explored: (1) \textbf{true}: Use the label distribution from the labeled set $L$. (2) \textbf{top-\textit{nk}}: Use the label distribution from the most confident $n \times k$ instances given by the prediction module.

Fig. \ref{fig::label_distribution_results} shows the DualRE model performance when different relation distributions are used. It can be concluded that the best performance is achieved at top-$3k$, whose distribution lies between true and top-$k$. Intuitively, the retrieved instances work better if they follows the true distribution. Due to the unbalanced distribution, however, the long-tail relations usually have fewer instances and are harder to learn. The true relation distribution forces the model to attend to these relations and selection quality goes down. On the other hand, top-$k$ distribution is too biased towards frequent relations. Retrieving more of these instances does not add much information for models to learn. Therefore, Top-$3k$ distribution can be regarded as a combination of above two strategies and works better.


\smallskip \noindent \textbf{Quality of Retrieved Instances}. 
To investigate the underlying reason for performance gain in DualRE, we conduct studies to investigate the hypotheses we have. A natural hypothesis is to attribute the performance gain mainly to quality of retrieved instances. It is under discussion how this ``quality'' can be defined as an objective metric. In our study, we regard quality as the precision of retrieved instances. We can technically evaluate the precision because we know the ground-truth labels of the ``unlabeled data''.

\begin{figure}
	\centering
	\subfigure{\includegraphics[width=0.48\linewidth]{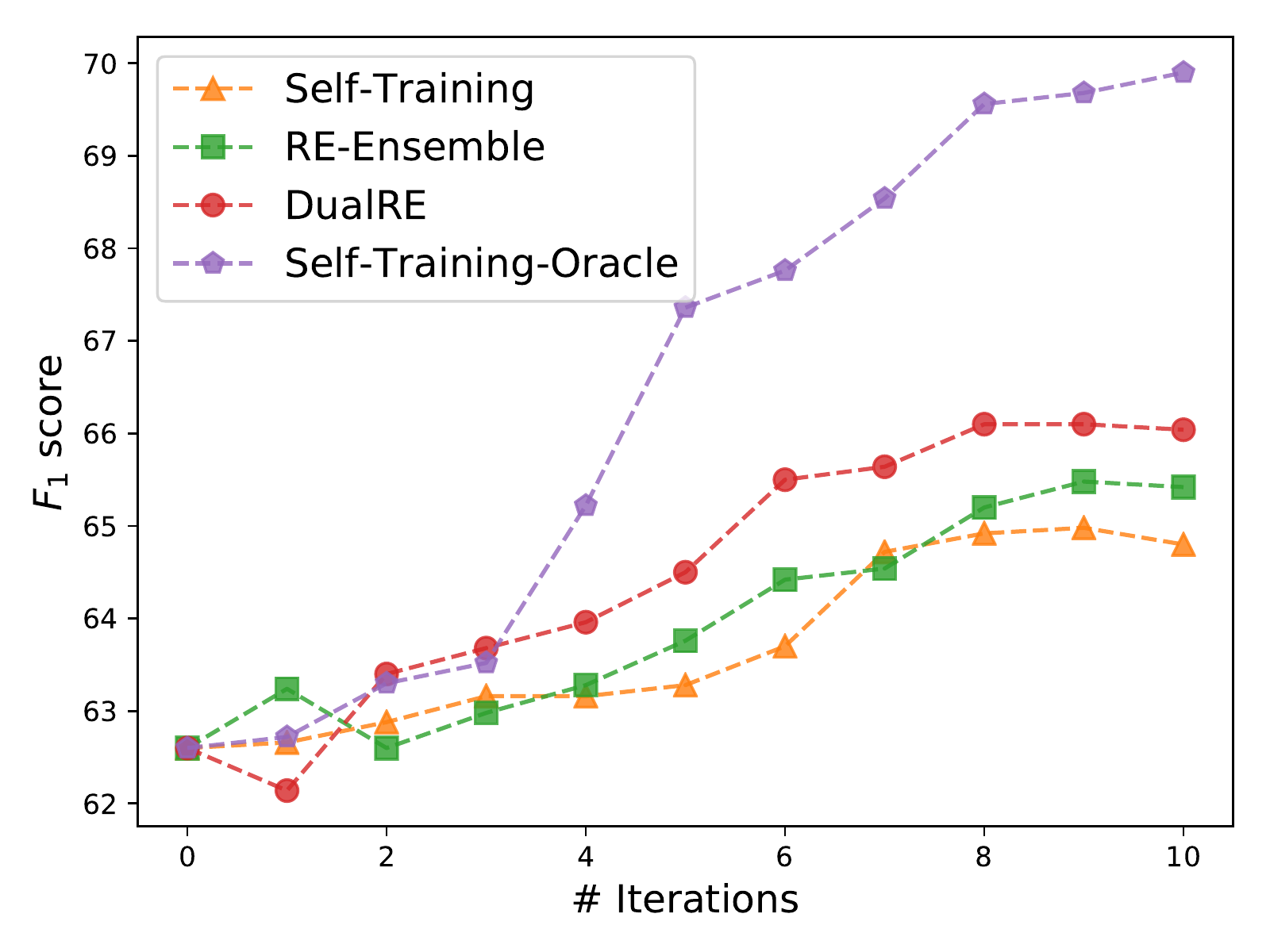}}
	\subfigure{\includegraphics[width=0.48\linewidth]{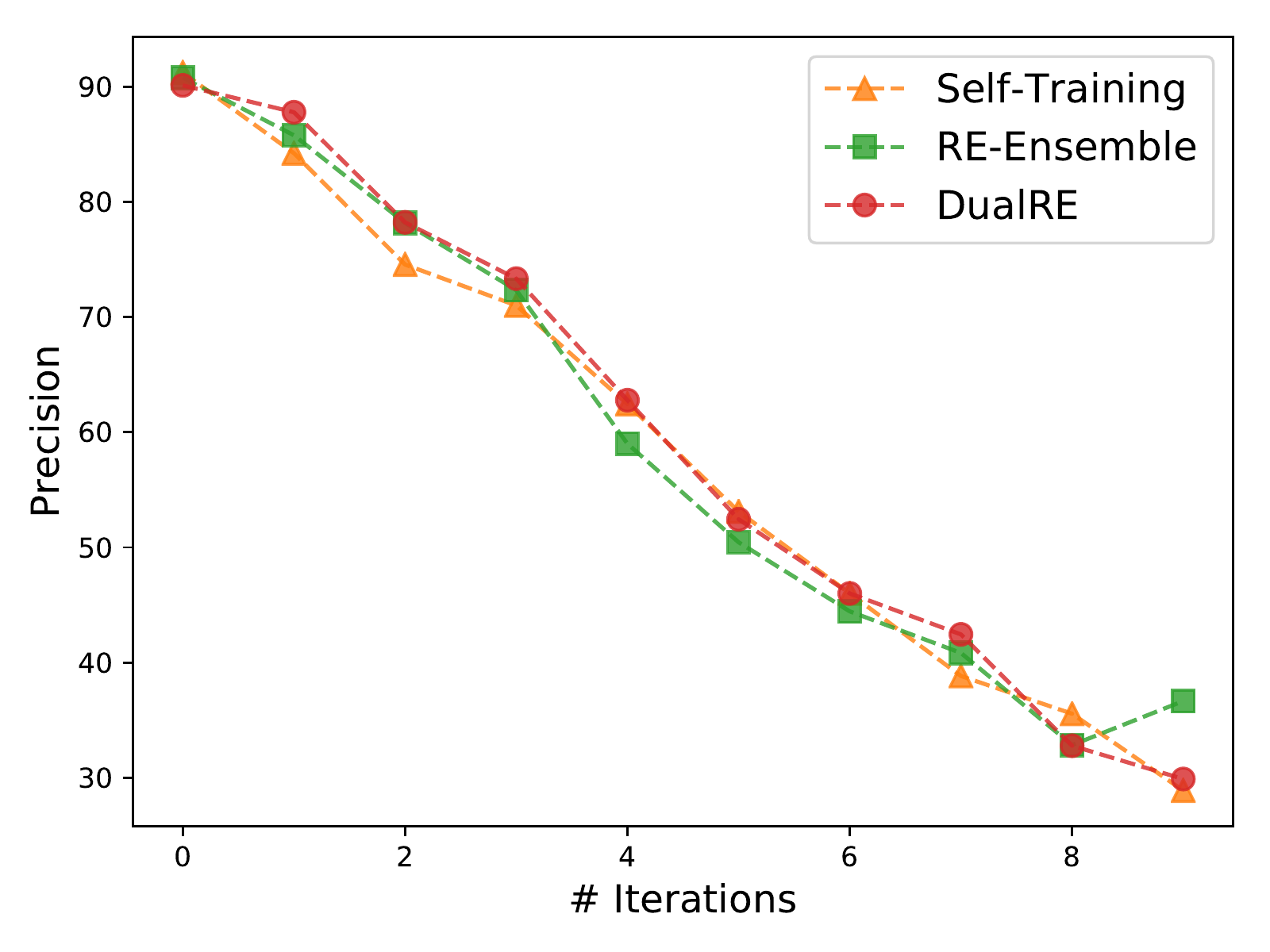}}
	\caption{\textbf{Analysis on Quality of Retrieved Instances.} \textbf{Left:} Convergence curve of test $F_1$ score for different methods; \textbf{Right:} Precision of the relation mentions selected in each iteration. Both are evaluated with 10\% labeled data and 50\% unlabeled data on the SemEval dataset.}
	\label{fig::iteration_results}
	\vspace{-1.0cm}
\end{figure}

We designed the following experiments to investigate this hypothesis. For semi-supervised models that are trained in an iterative method (Self-Training, RE-Ensemble and DualRE), we evaluate their $F_1$ scores on the test set in each iteration together with the precision of its promoted pseudo-labeled instances.

To get a better idea of the upper bound performance gain in this setting, we construct an oracle model: ``Self-Training-Oracle''. Similar to Self-Training, we rank the instances by their probability towards predicted label and select new instances in each iteration. But here we assume that the model can perfectly select those correctly pseudo-labeled instances and add them to the training set in each iteration. If the number of selected instances is smaller than expected, we pick from the rest of the samples in order. 

Experimental results are shown in Fig. \ref{fig::iteration_results}. Comparing results on DualRE and other semi-supervised models, the precision for retrieved instances in each iteration are generally higher in DualRE, which indicates that DualRE can select instances with better quality. This directly contributes to the performance gain in $F_1$ score in each iteration. To measure the upper bound of semi-supervised models that fall in the self-training category, the performance of Self-Training-Oracle indicates that the potential for such models are high enough to be comparable with RE-Gold (refer to Table~\ref{tab::semeval_results}). But this upper bound is extremely hard to approach because the model does not have additional supervision to guide the instance selection process.

%% file: 2-related-work.tex
\section{Related Work}
\label{sec::related_work}

\smallskip 
\noindent \textbf{Relation extraction.}
Our work is related to sentence-level relation extraction, which aims at predicting the relation label between two entities in a sentence. Traditional relation extraction methods~\cite{agichtein2000snowball,qu2017automatic} seek to find discriminative textual patterns with some labeled sentences, and further leverage these patterns for relation extraction. Nevertheless, the learned patterns may have difficulties matching with unseen sentences, as patterns are composed of discrete words. Recent studies leverage deep neural networks such as Long-Short Term Memory Networks (LSTM)~\cite{hochreiter1997long} and Convolutional Neural Networks (CNN)~\cite{zeng2015distant} for relation extraction, achieving much better performance. However, all these methods are fully-supervised, which require a large amount of labeled data to train effective models. By contrast, our proposed approach is semi-supervised, which requires much fewer labeled sentences during training.

\smallskip 
\noindent \textbf{Semi-supervised learning.}
In the literature of semi-supervised learning, there are some methods which can be applied to our problem. One major category is the self-ensembling method~\cite{french2017self}, which assumes that the predicted labels of the unlabeled instances should remain unchanged under small perturbations on the instances or the model parameters. In this way, they guarantee the local consistency of the prediction and hence improve the performance. However, when applied to our problem, such a training framework relies solely on local consistency as supervision signal, which is weak compared to labeled data, therefore the improvement is marginal. Another category is the self-training method~\cite{rosenberg2005semi}, which iteratively promotes instances from unlabeled data. Self-training methods impose a stronger supervision signal and generally achieve better performance. However, it suffers from semantic drift \cite{curran2007minimising}, which accumulates systematic errors from prediction model in each iteration. Co-training \cite{blum1998combining} is an attempt to eliminate this problem. It promotes unlabeled data using two models that are trained on different views of data, which essentially relies on the assumptions that two different sets of features are independent and sufficient for prediction. This assumption is not easy to follow in relation extraction task, in which relation mention is the only source of input. Our DualRE model, instead, introduces a dual retrieval module to correct the bias from prediction module. Therefore we are able to collect extra training data with higher quality to avoid semantic drifting.

\smallskip 

\noindent \textbf{Dual learning.}
Our work is also related to dual learning~\cite{he2016dual,xia2017dual,tang2017question}, which can be generalized to the concept of jointly learning the primal and dual models to boost the performance of each other. For example, English-French translation with French-English translation~\cite{he2016dual}, image classification with image generation~\cite{xia2017dual}, question answering and question generation~\cite{tang2017question} can be treated together to leverage the duality of two tasks. Different from their supervised settings, we apply the concept of dual model in semi-supervised settings and propose an algorithm that effectively utilize the unlabeled data. We are also the first to perform dual model learning in the relation extraction task.

%% file: 6-conclusion.tex
\section{Conclusion}

In this paper, we studied the problem of relation extraction in the semi-supervised setting, where a few labeled data and an auxiliary unlabeled corpus are provided. We proposed a principled framework consisting of a prediction module and a retrieval module, which focus on the primal prediction task and the dual retrieval task respectively. During training, we encouraged them to mutually enhance each other through annotating unlabeled sentences as additional training data. A joint optimization method was proposed for optimizing our framework, which alternates between updating the prediction module and updating the retrieval module. Experimental results on two public datasets proved the effectiveness of our proposed framework. Interesting future work directions include switching the retrieval module to text generative models, extending our framework to deal with various text classification tasks, and studying the connection to variational auto-encoder method.

\section{Acknowledgement}

This work has been supported in part by National Science Foundation SMA 18-29268, Amazon Faculty Award, and JP Morgan AI Research Award. We would like to thank all the collaborators in INK research lab for their constructive feedback on the work.